\documentclass[11pt]{article}

\usepackage[final]{acl}

\usepackage{times}
\usepackage{latexsym}

\usepackage[T1]{fontenc}

\usepackage[utf8]{inputenc}

\usepackage{microtype}

\usepackage{inconsolata}

\usepackage{graphicx}

\usepackage{colortbl}
\usepackage{booktabs}
\usepackage{subcaption}
\usepackage{amsmath}
\usepackage{amssymb}
\usepackage{algorithm}
\usepackage{algorithmic}
\usepackage[dvipsnames]{xcolor}
\usepackage{tabularx}
\usepackage{array}
\usepackage{multirow}
\usepackage{multirow}
\usepackage{titletoc}

\title{How Reliable Is Visual Token Selection? \\Rethinking Pruning in Multimodal LLMs}

\author{Chaehee Song\quad Seokil Ham\quad KiHyun Seol\quad Jihyun Lee\quad Yubin Choi\quad Changick Kim\\
        Korea Advanced Institute of Science and Technology (KAIST) \\
      \texttt{\{chaehee.song,gkatjrdlf,seollab,jihyun,choibinbin,changick\}@kaist.ac.kr}}

\begin{document}
\maketitle
\begin{abstract}
Visual token pruning improves the efficiency of multimodal large language models (MLLMs) by removing redundant visual tokens, yet the reliability of existing selection signals has not been systematically examined. We analyze token selection under semantics-preserving perturbations and across representative pruning criteria, revealing two limitations. First, attention-based importance shifts non-uniformly under perturbations, so magnitude alone cannot distinguish stable evidence from unstable saliency (\emph{local instability}). Second, attention- and diversity-based methods select substantially different subsets despite comparable performance, showing that useful evidence is not confined to a unique subset (\emph{global inconsistency}). These findings motivate \textbf{reliable} token-set construction. We instantiate this perspective in \textbf{CoRePruner}, a training-free framework combining instruction-aware diversity with reliability-aware attention. Across image benchmarks, multiple MLLM architectures, and token budgets, CoRePruner improves performance, especially under aggressive pruning, and remains competitive on video benchmarks. Our results establish reliability as a central consideration in visual token selection.
\end{abstract}

\section{Introduction}
\label{sec:intro}

\begin{figure}[t]
    \centering
    \includegraphics[width=1.0\linewidth]{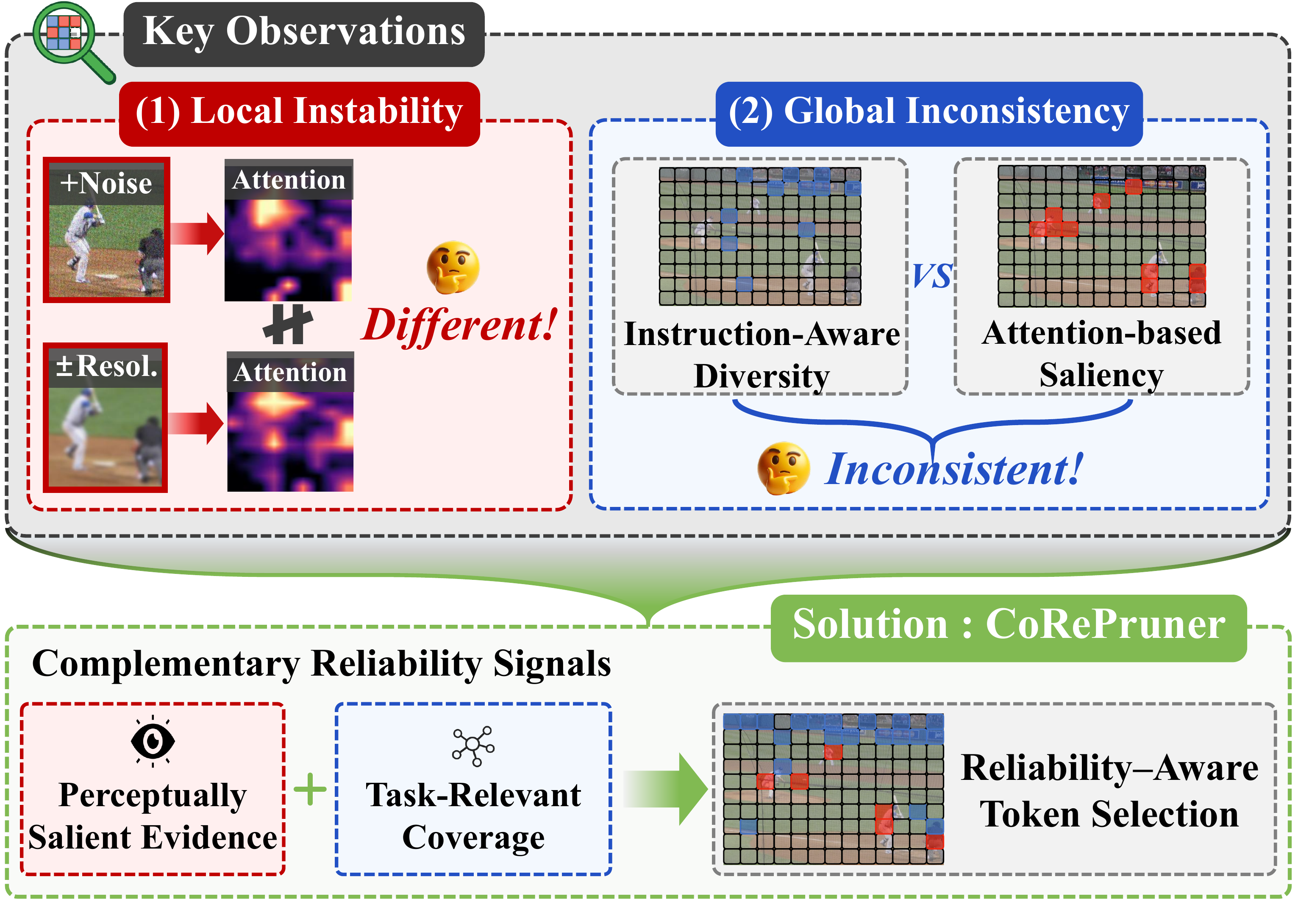}
    \caption{
    \textbf{Limitations of existing criteria and CoRePruner.} (1) \emph{Local instability}: attention shifts across semantically equivalent inputs. (2) \emph{Global inconsistency}: attention and diversity select different tokens yet perform similarly. CoRePruner combines stable attention with instruction-aligned coverage to construct reliable token sets.}
    \label{fig:concept}
\end{figure}

Recent advances in Multimodal Large Language Models (MLLMs)~\cite{liu2023visual, li2023blip, team2024gemini, liu2024llavanext, bai2025qwen3} have substantially improved multimodal understanding and visual reasoning. These models increasingly process high-resolution images and videos, producing hundreds or even thousands of visual tokens per input. Yet many visual tokens carry redundant or limited semantic information, and the resulting mismatch between computational cost and semantic utility makes visual token reduction a key challenge for practical deployment.

Existing pruning methods estimate token importance using attention~\cite{chen2024image,xing2024pyramiddrop,ye2025fit}, feature diversity~\cite{zhang2026beyond,alvar2025divprune}, or both~\cite{zhang2025beyond,deng2026scope}. 
Despite achieving strong efficiency--performance trade-offs, prior work has largely focused on refining or combining these signals rather than systematically examining their reliability across inputs and criteria.

\begin{figure*}[t]
    \centering
    \begin{minipage}[t]{0.50\textwidth}
        \vspace{0pt}
        \centering
        \includegraphics[width=\linewidth]{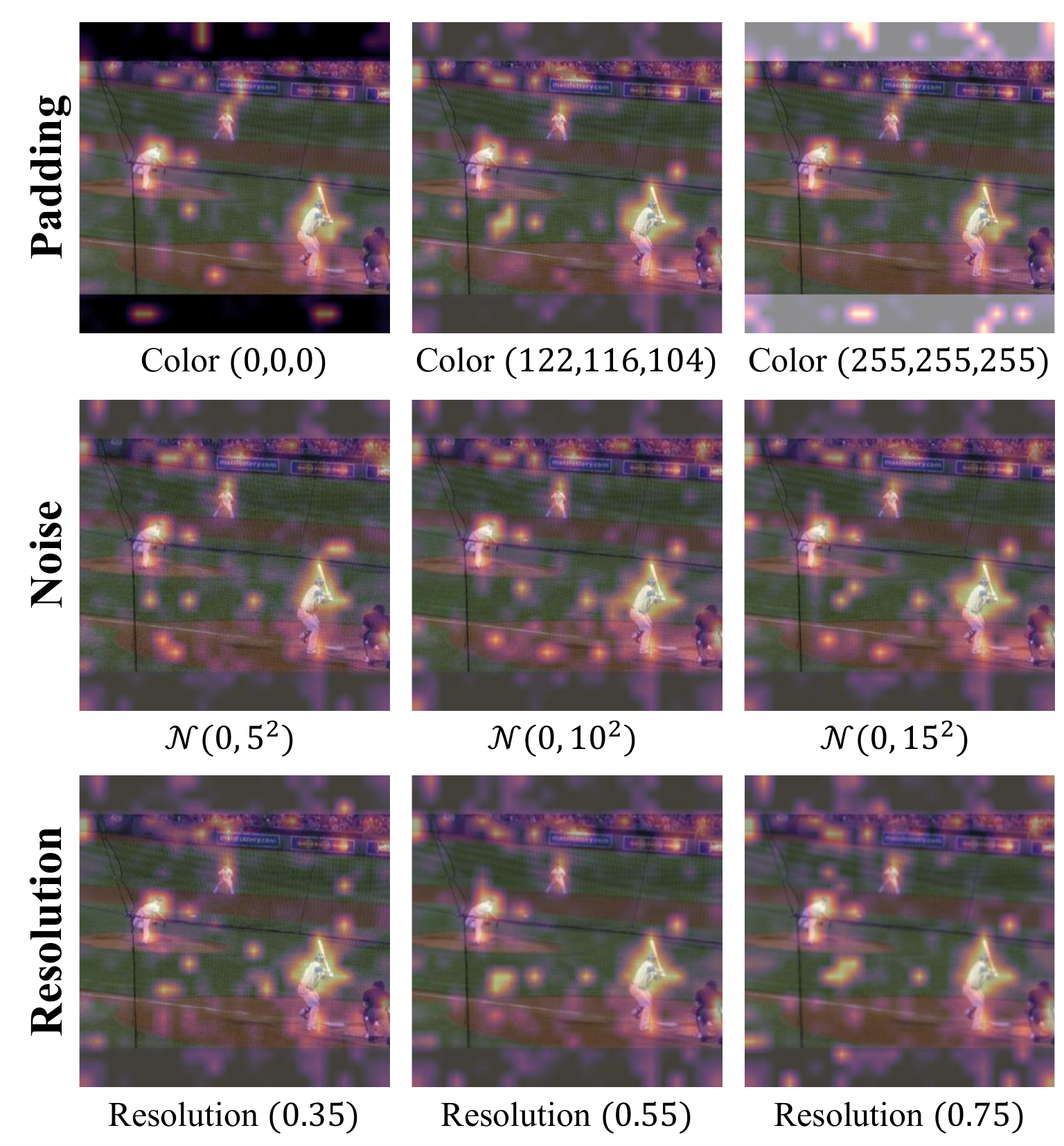}
        \caption{
            Attention instability under semantics-preserving perturbations. 
            Perturbation parameters are shown in parentheses: padding colors (RGB), Gaussian noise (std), and resolution scaling (ratio).
        }
        \label{fig:attention-vis}
    \end{minipage}%
    \hspace{0.02\textwidth}
    \begin{minipage}[t]{0.45\textwidth}
        \vspace{0pt} 
        \centering
        \includegraphics[
            width=\linewidth
        ]{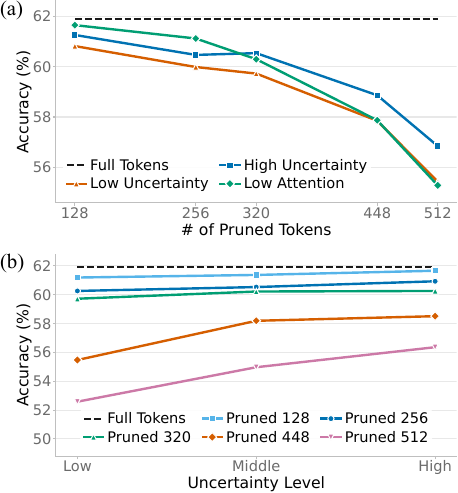}
        \vspace{-1mm}
        \caption{
            Effect of attention uncertainty on pruning utility (GQA).
            (a) Accuracy after removing high-uncertainty, low-uncertainty, or low-attention tokens.
            (b) Accuracy after removing low-, middle-, or high-uncertainty tokens under different pruning budgets.
        }
        \label{fig:pruning-uncertainty}
    \end{minipage}
\end{figure*}

We investigate this issue by analyzing token selection under semantics-preserving perturbations and across representative pruning criteria. Our analysis reveals two limitations of existing selection criteria (Figure~\ref{fig:concept}). First, attention shifts markedly under such perturbations, but not uniformly: tokens on task-relevant regions tend to remain consistently attended, whereas less informative regions fluctuate and can transiently receive high attention. Attention magnitude alone therefore cannot distinguish stable evidence from unstable saliency (\emph{local instability}). Second, attention- and diversity-based methods select substantially different token subsets despite comparable downstream performance. This suggests that useful evidence is not confined to a unique subset and that different criteria preserve distinct forms of information (\emph{global inconsistency}).

Together, these findings show that token selection is locally unreliable within criteria and globally inconsistent across them. This motivates moving beyond merely identifying important tokens toward constructing \textbf{reliable} token sets that preserve stable and complementary visual evidence.

Guided by these findings, we introduce \textbf{CoRePruner} (\textbf{Co}mplementary \textbf{Re}liability-aware Token Pruning), a training-free, plug-and-play framework for reliable token-set construction. It combines
instruction-aware diversity for contextual coverage with attention anchors
scored by magnitude and stability. Extensive experiments across multiple benchmarks and architectures show consistent gains over prior pruning methods, especially under aggressive token budgets. These results support our central claim that reliability should guide the design and evaluation of visual token pruning.

\section{Related Work}
\subsection{Multimodal Large Language Models}

Building on advances in LLMs~\cite{achiam2023gpt, bai2023qwen, liu2025deepseek}, MLLMs integrate visual representations into language backbones through query-based compression~\cite{dai2023instructblip} or by projecting CLIP patch tokens into the language model~\cite{radford2021learning, liu2024improved}. Recent models support larger visual inputs through dynamic tiling~\cite{liu2024llavanext}, adaptive patch merging~\cite{Qwen2.5-VL, bai2025qwen3}, or sampled video frames~\cite{lin2024video, maaz2024video}. These designs produce hundreds to thousands of visual tokens, motivating visual token reduction.

\begin{figure*}[t]
    \centering
    \begin{minipage}[t]{0.35\textwidth}
        \vspace{0pt}
        \centering
        \footnotesize
            \footnotesize
            \setlength{\tabcolsep}{3pt}
            \begin{tabular}{l|cc|c}
                \toprule
                \textbf{Benchmark} & \textbf{Attention} & \textbf{Diversity} & \textbf{IoU} \\
                \midrule
                \rowcolor{gray!20}
                \multicolumn{4}{c}{\textit{Retain 128 Tokens \textcolor{green!60!black}{($\downarrow$ 77.8\%)}}} \\
                \midrule
                VQA\textsuperscript{Text} & 52.8 & 52.6 & 0.39 \\
                VQA\textsuperscript{V2} & 76.2 & 76.6 & 0.34 \\
                MMB\textsuperscript{EN} & 62.5 & 63.1 & 0.39 \\
                GQA & 58.5 & 59.8 & 0.30 \\
                \midrule
                \rowcolor{gray!20}
                \multicolumn{4}{c}{\textit{Retain 64 Tokens \textcolor{green!60!black}{($\downarrow$ 88.9\%)}}} \\
                \midrule
                VQA\textsuperscript{Text} & 52.1 & 51.7 & 0.26 \\
                VQA\textsuperscript{V2} & 73.2 & 75.4 & 0.20 \\
                MMB\textsuperscript{EN} & 61.3 & 61.2 & 0.22 \\
                GQA & 55.6 & 58.7 & 0.18 \\
                \bottomrule
            \end{tabular}
        \captionof{table}{
            Attention- and diversity-based pruning achieve broadly comparable accuracy despite low token-set overlap; the overlap decreases under tighter budgets.
        }
        \label{tab:attention_vs_diversity}
    \end{minipage}%
    \hfill
    \begin{minipage}[t]{0.63\textwidth}
        \vspace{0pt}
        \centering
        \includegraphics[width=1.0\columnwidth]{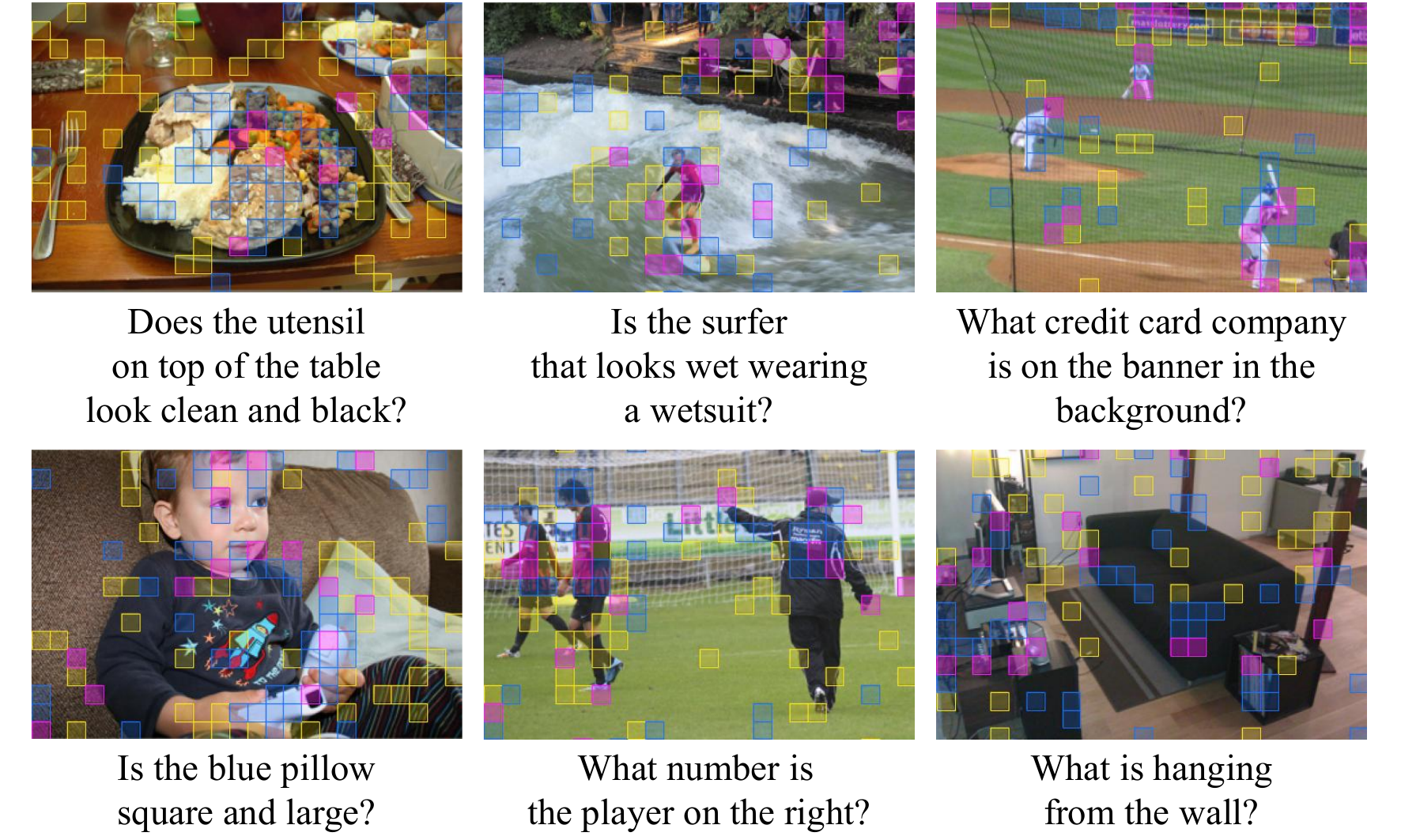}
        \caption{Overlap between attention- and diversity-based selections: \textcolor{RoyalBlue}{attention-only (blue)}, \textcolor{Goldenrod}{diversity-only (yellow)}, and \textcolor{VioletRed}{shared tokens (pink)}.}
        \label{fig:iou_visualize}
    \end{minipage}
\end{figure*}

\subsection{Token Pruning in Multimodal Large Language Models}
Existing pruning methods select visual tokens using attention-based saliency, feature diversity, or both. Attention-based methods use LLM-side attention to prune tokens within language-model layers (FastV~\cite{chen2024image}, SparseVLM~\cite{zhang2024sparsevlm}), whereas VisionZip~\cite{yang2025visionzip} uses vision-encoder attention to select dominant tokens and merges the remaining tokens by similarity. 
Diversity-based methods promote feature-space coverage through Max--Min selection (DivPrune~\cite{alvar2025divprune}) or instruction-aware determinantal point processes (CDPruner~\cite{zhang2026beyond}). 
Hybrid methods combine saliency with redundancy reduction or coverage (VisPruner~\cite{zhang2025beyond}, SCOPE~\cite{deng2026scope}), while EADP~\cite{wang2026combating} combines entropy-filtered instruction relevance with spatially structured subset selection.
Despite their different objectives, these methods largely focus on refining or combining selection signals without systematically examining signal reliability. We examine this gap and identify two limitations of existing selection criteria: \emph{local instability} and \emph{global inconsistency}. These findings motivate recasting visual token pruning as reliable token-set construction.  
\section{Findings} 
\label{sec: observation}

We test the reliability of existing selection criteria along two axes: attention stability under semantics-preserving perturbations and subset consistency across pruning criteria. Experimental details are provided in Appendices~\ref{sec:app-observation1},
\ref{sec:app-observation2}, and~\ref{sec:app-perturbations}.

\subsection{Finding 1: Attention-Based Importance Is Locally Unstable}
\label{sec:observation1}
Attention is widely used as a proxy for token importance, assuming highly attended tokens are more useful. 
Figure~\ref{fig:attention-vis} shows that semantics-preserving perturbations---padding-color changes, mild noise injection, and resolution scaling---markedly alter attention distributions and token rankings. The changes are not uniform: tokens on visually meaningful regions (e.g., players, ball, and banners) remain consistently attended, whereas less informative regions fluctuate and can transiently receive high attention. We therefore test whether stability predicts pruning-induced accuracy loss among tokens with comparable mean attention.

Stability is quantified by the coefficient of variation across $K$ perturbed inputs. For the $i$-th visual token with attention scores $\{a_i^k\}_{k=1}^{K}$,
\begin{equation}
\label{eq:uncertainty}
    u_i = \frac{\sigma(\{a_i^k\}_{k=1}^{K})}{\mu(\{a_i^k\}_{k=1}^{K})+\epsilon},
\end{equation}
where $\mu$ and $\sigma$ denote the mean and standard deviation, respectively, and $\epsilon=10^{-12}$ is a small constant. A larger $u_i$ indicates less stable attention.

Removing high-uncertainty tokens preserves higher accuracy than removing low-uncertainty tokens (Figure~\ref{fig:pruning-uncertainty}(a)). The gap widens under tighter budgets, and high-uncertainty pruning even outperforms low-attention pruning at tighter budgets. At fixed budgets, accuracy increases monotonically from low- to middle- to high-uncertainty removal (Figure~\ref{fig:pruning-uncertainty}(b)).

Attention magnitude alone is therefore insufficient for reliable token selection: tokens with similar attention can differ substantially in pruning utility, and stability distinguishes them. We refer to this failure as \emph{local instability}.

\begin{figure*}[t!]
    \centering
    \includegraphics[width=1.0\linewidth]{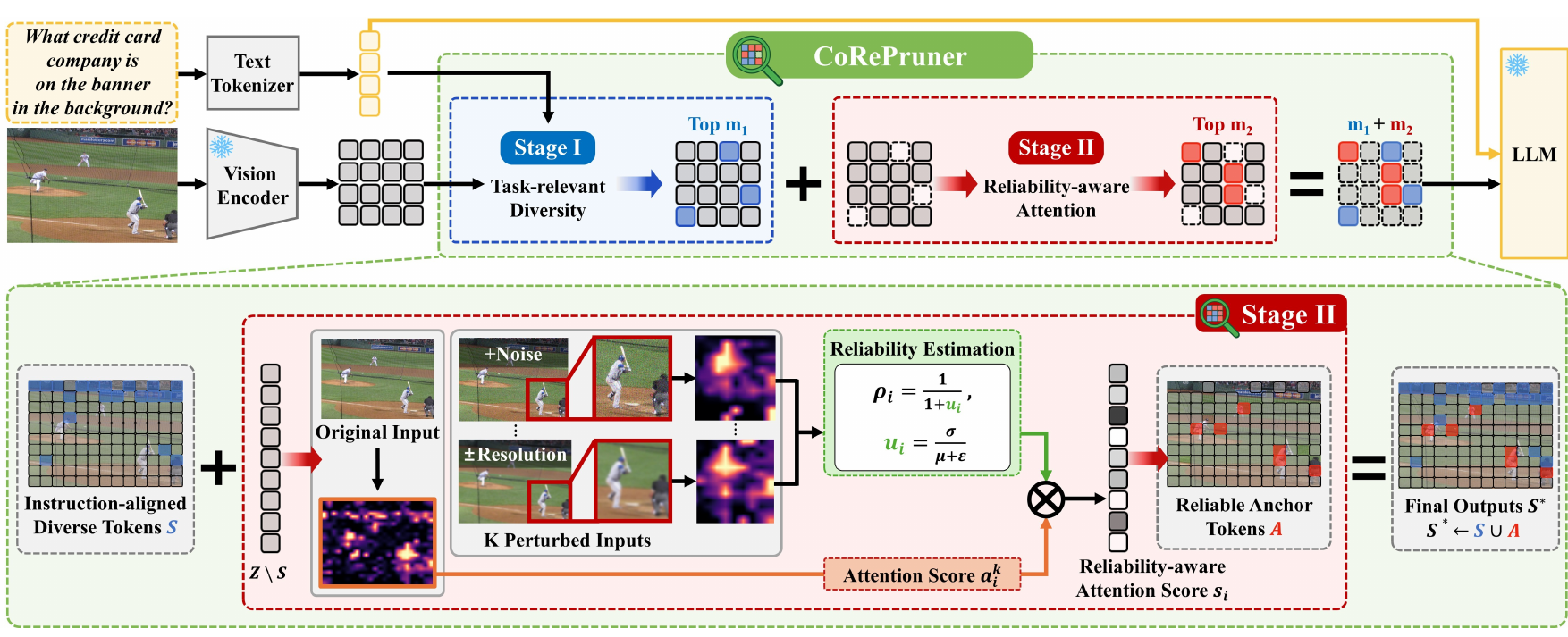}
    \caption{
        Method overview of CoRePruner.
        Stage~I selects task-relevant diverse tokens using DPP with an instruction-modulated kernel. Stage~II selects reliability-aware attention anchors from the remaining tokens. The final token set is $S^* = S \cup A$.
    }
    \label{fig:overview}
\end{figure*}

\subsection{Finding 2: Pruning Criteria Are Globally Inconsistent}\label{sec:observation2}

We examine whether pruning criteria identify the same informative tokens by comparing instruction-aware diversity~\cite{zhang2026beyond} and CLS-token attention~\cite{zhang2025beyond}, representative diversity- and attention-based criteria. We isolate the criteria rather than the full pipelines so that the comparison contrasts selection signals directly.

Table~\ref{tab:attention_vs_diversity} shows that both strategies perform comparably across tasks, yet the intersection-over-union (IoU) between their selected token sets is only 0.30--0.39 at 128 tokens and drops to 0.18--0.26 at 64 tokens.
Figure~\ref{fig:iou_visualize} shows the same pattern qualitatively: the two criteria retain largely distinct image regions. Instruction-aware diversity preserves distributed regions aligned with the input instruction, whereas CLS-token attention emphasizes visually salient objects.

Comparable performance from largely disjoint subsets suggests that useful visual evidence is not confined to a unique token set, but can be preserved by multiple, different subsets. Low overlap alone, however, does not indicate that either criterion is unreliable, nor that the two criteria are complementary; it only shows that they select different tokens. We therefore examine complementarity at the case level. On GQA with 64 tokens, the diversity-only and attention-only variants of our method (Section~\ref{sec:method}) reach 58.7 and 56.0, yet each answers 8.1\% and 5.3\% of questions that the other misses, i.e., they disagree in correctness on 13.4\% of questions. In contrast, two attention-only runs with different perturbation sets disagree on only 2.4\%, a $5.6\times$ gap, so the difference cannot be attributed to run-to-run variation. Different criteria therefore retain distinct yet useful evidence, and refining a single criterion may overlook evidence captured by another. We refer to this cross-criterion disagreement as \emph{global inconsistency}. Unlike local instability (Section~\ref{sec:observation1}), which concerns how stable a single criterion is under input perturbations, global inconsistency concerns the complementarity between criteria. 

\begin{table*}[!h]
\centering
\footnotesize
\setlength{\tabcolsep}{2.3pt} 
\renewcommand{\arraystretch}{1.02} 
\begin{tabular*}{\textwidth}{@{\extracolsep{\fill}}l*{9}{c}cc@{}}
\hline
\textbf{Method} & \textbf{VQA}\textsuperscript{v2} & \textbf{GQA} & \textbf{VQA}\textsuperscript{Text} & \textbf{MME} &
\textbf{MMB}\textsuperscript{EN} & \textbf{MMB}\textsuperscript{CN} & \textbf{MMVet} & \textbf{SQA}\textsuperscript{Img} & \textbf{POPE} & \textbf{Avg.} & \textbf{Rel.} \\
\hline

\rowcolor{gray!20}
\multicolumn{12}{c}{\it{Upper Bound, All 576 Tokens (100\%)}}\\
\hline
LLaVA-1.5-7B & 78.5 & 61.9 & 58.2 & 1506.5 & 64.7 & 58.1 & 31.1 & 69.5& 85.9 & 64.8 & 100\%\\
\hline

\rowcolor{gray!20}
\multicolumn{12}{c}{\it{Retain 128 Tokens \textcolor{green!60!black}{($\downarrow$ 77.8\%)}}}\\
\hline
FastV {\scriptsize\ensuremath{(\mathrm{ECCV24})}}        & 61.8 & 49.6 & 50.6 & 1368.9 & 56.1 & 51.4 & 28.1 & 60.2 & 59.6 & 54.0 & 84.2\% \\
SparseVLM {\scriptsize\ensuremath{(\mathrm{ICML25})}}     & 73.8 & 56.0 & 54.9 & 1399.3 & 60.0 & 51.1 & 30.0 & 67.1 & 80.5 & 60.4 & 93.2\% \\
PDrop {\scriptsize\ensuremath{(\mathrm{CVPR25})}}        & 74.3 & 57.1 & 56.6 & 1444.1 & 61.6 & 55.3 & 27.6 & 68.4 & 77.5 & 61.2 & 94.2\% \\
VisionZip {\scriptsize\ensuremath{(\mathrm{CVPR25})}}    & 75.6 & 57.6 & 56.8 & 1436.9 & 62.0 & 56.7 & 32.6 & 68.9 & 83.2 & 62.8 & 97.4\% \\
DivPrune {\scriptsize\ensuremath{(\mathrm{CVPR25})}}     & 76.0 & 59.4 & 55.9 & 1405.1 & 61.5 & 54.8 & 30.6 & 68.6 & 87.0 & 62.7 & 96.6\% \\
VisPruner {\scriptsize\ensuremath{(\mathrm{ICCV25})}}    & 75.8 & 58.2 & 57.0 & \textbf{1461.4} & 62.7 & \textbf{57.3} & \textbf{33.7} & \textbf{69.1} & 84.6 & 63.5 & 98.6\% \\
CDPruner {\scriptsize\ensuremath{(\mathrm{NeurIPS25})}}  & 76.6 & 59.9 & 56.2 & 1431.4 & \underbar{63.1} & 55.0 & 32.8 & \underbar{69.0} & \textbf{87.7} & 63.5 & 98.3\% \\
EADP {\scriptsize\ensuremath{(\mathrm{ECCV26})}}
& \underbar{76.7} & 60.0 & 56.5 & 1439.0
& 62.7 & 55.2 & 32.5 & \underbar{69.0}
& 87.2 & 63.5 & 98.3\% \\\midrule
\textbf{CoRePruner} {\scriptsize\ensuremath{(\mathrm{Ours})}}    & \textbf{76.9} & \textbf{60.3} & \underbar{57.3} & \underbar{1458.4} & \textbf{63.2} & 56.9 & \underbar{33.2} & 68.7 & \underbar{87.6} & \textbf{64.1} & \textbf{99.3\%} \\
\textbf{CoRePruner$^{\dagger}$} {\scriptsize\ensuremath{(\mathrm{Ours})}}    & \textbf{76.9} & \underbar{60.2} & \underbar{57.3} & 1449.4 & 63.0 & 56.9 & 32.8 & 68.7 & 87.4 & \underbar{64.0} & \underbar{99.0\%} \\
\textbf{CoRePruner$^{\ddagger}$} {\scriptsize\ensuremath{(\mathrm{Ours})}}    & \textbf{76.9} & \textbf{60.3} & \textbf{57.6} & 1449.7 & 63.0 & \underbar{57.0} & 33.1 & 68.6 & \underbar{87.6} & \textbf{64.1} & \textbf{99.3\%} \\
\hline

\rowcolor{gray!20}
\multicolumn{12}{c}{\it{Retain 64 Tokens \textcolor{green!60!black}{($\downarrow$ 88.9\%)}}}\\
\hline
FastV {\scriptsize\ensuremath{(\mathrm{ECCV24})}}        & 55.0 & 46.1 & 47.8 & 973.5 & 48.0 & 42.7 & 25.8 & 51.1 & 48.0 & 45.9 & 72.4\%  \\
SparseVLM {\scriptsize\ensuremath{(\mathrm{ICML25})}}     & 68.2 & 52.7 & 51.8 & 1190.4 & 56.2 & 46.1 & 23.3 & 62.2 & 75.1 & 55.0 & 84.2\%  \\
PDrop {\scriptsize\ensuremath{(\mathrm{CVPR25})}}        & 56.3 & 47.5 & 50.6 & 982.2 & 58.8 & 36.6 & 17.7 & \underbar{69.0} & 40.8 & 47.4 & 73.1\% \\
VisionZip {\scriptsize\ensuremath{(\mathrm{CVPR25})}}    & 72.4 & 55.1 & 55.5 & 1365.2 & 60.1 & 55.4 & 31.7 & \underbar{69.0} & 77.0 & 60.5 & 94.0\% \\
DivPrune {\scriptsize\ensuremath{(\mathrm{CVPR25})}}     & 74.1 & 57.5 & 54.5 & 1334.7 & 60.1 & 52.3 & 28.1 & 68.0 & 85.8 & 60.8 & 93.4\% \\
VisPruner {\scriptsize\ensuremath{(\mathrm{ICCV25})}}    & 72.7 & 55.4 & \underbar{55.8} & 1369.9 & 61.3 & 55.1 & \textbf{32.3} & \textbf{69.1} & 80.4 & 61.2 & 95.0\% \\
CDPruner {\scriptsize\ensuremath{(\mathrm{NeurIPS25})}}  & 75.4 & 58.6 & 55.3 & 1415.1 & 61.1 & 53.2 & 30.5 & 68.1 & \underbar{87.5} & 62.3 & 96.0\% \\
EADP {\scriptsize\ensuremath{(\mathrm{ECCV26})}}
& \underbar{75.6} & \textbf{59.4} & 55.0 & 1403.6
& \textbf{61.9} & 53.1 & 29.4 & 68.9
& 86.5 & 62.2 & 95.7\% \\\midrule
\textbf{CoRePruner} {\scriptsize\ensuremath{(\mathrm{Ours})}}    & \textbf{75.8} & 59.0 & \textbf{56.4} & 1438.2 & 61.4 & \textbf{56.5} & 31.4 & 68.9 & \textbf{87.6} & \underbar{63.2} & \underbar{97.6\%} \\
\textbf{CoRePruner$^{\dagger}$} {\scriptsize\ensuremath{(\mathrm{Ours})}}    & \textbf{75.8} & 59.0 & \textbf{56.4} & \underbar{1438.6} & \underbar{61.7} & \underbar{56.4} & \underbar{31.6} & \underbar{69.0} & \underbar{87.5} & \textbf{63.3} & \textbf{97.7\%} \\
\textbf{CoRePruner$^{\ddagger}$} {\scriptsize\ensuremath{(\mathrm{Ours})}}    & \textbf{75.8} & \underbar{59.1} & \textbf{56.4} & \textbf{1445.7} & 61.5 & 55.9 & 30.7 & 68.5 & \underbar{87.5} & 63.1 & 97.3\% \\
\hline
\rowcolor{gray!20}
\multicolumn{12}{c}{\it{Retain 32 Tokens \textcolor{green!60!black}{($\downarrow$ 94.4\%)}}}\\
\hline
VisionZip {\scriptsize\ensuremath{(\mathrm{CVPR25})}}    & 67.1 & 51.8 & 53.1 & 1251.2 & 57.7 & 50.3 & 25.5 & 68.8 & 68.7 & 56.2 & 86.7\% \\
DivPrune {\scriptsize\ensuremath{(\mathrm{CVPR25})}}     & 71.2 & 54.9 & 52.9 & 1284.9 & 57.6 & 49.1 & 26.3 & 68.6 & 81.5 & 58.5 & 89.7\% \\
VisPruner {\scriptsize\ensuremath{(\mathrm{ICCV25})}}    & 67.7 & 52.2 & 53.9 & 1271.0 & 58.4 & 52.7 & 28.8 & 69.2 & 72.7 & 57.7 & 89.5\% \\
CDPruner {\scriptsize\ensuremath{(\mathrm{NeurIPS25})}}  & \underbar{73.6} & 57.0 & 53.2 & 1373.0 & 59.6 & 49.6 & 27.8 & \textbf{69.5} & \underbar{87.9} & 60.8 & 93.1\% \\ 
EADP {\scriptsize\ensuremath{(\mathrm{ECCV26})}}
& \textbf{74.0} & \textbf{58.2} & 52.7 & 1347.0
& 59.4 & 50.1 & \textbf{30.6} & 69.3
& 86.7 & 60.9 & 93.9\% \\\midrule
\textbf{CoRePruner} {\scriptsize\ensuremath{(\mathrm{Ours})}}    & \textbf{74.0} & \underbar{57.1} & \textbf{55.0} & \underbar{1380.6} & \underbar{60.2} & \textbf{53.3} & 29.3 & \underbar{69.4} & \textbf{88.0} & \textbf{61.7} & \textbf{95.0\%} \\
\textbf{CoRePruner$^{\dagger}$} {\scriptsize\ensuremath{(\mathrm{Ours})}}    & \textbf{74.0} & \underbar{57.1} & \underbar{54.9} & \textbf{1389.5} & \underbar{60.2} & \underbar{53.1} & 28.4 & 69.3 & 87.7 & \underbar{61.6} & \underbar{94.5\%} \\
\textbf{CoRePruner$^{\ddagger}$} {\scriptsize\ensuremath{(\mathrm{Ours})}}    & \textbf{74.0} & 57.0 & \textbf{55.0} & 1377.6 & \textbf{60.7} & \textbf{53.3} & \underbar{29.5} & \underbar{69.4} & 87.8 & \textbf{61.7} & \textbf{95.0\%} \\
\hline
\end{tabular*}
\caption{
Performance comparison on LLaVA-1.5-7B under strict visual token budgets (128, 64, 32). CoRePruner, CoRePruner$^{\dagger}$, and CoRePruner$^{\ddagger}$ use cosine, RBF, and polynomial diversity kernels, respectively, demonstrating the plug-and-play nature. \textbf{Avg.} is the mean over 9 benchmarks, and \textbf{Rel.} is the average performance retention. CoRePruner achieves the best overall performance, with larger gains under aggressive pruning.
}
\label{tab:llava1.5}   
\end{table*}

\section{CoRePruner}
\label{sec:method}

Our findings motivate reliable token-set construction that preserves complementary evidence while favoring perturbation-stable signals. As a concrete instantiation, we introduce CoRePruner (\textbf{Co}mplementary \textbf{Re}liability-aware Token Pruning), a training-free framework that combines instruction-aligned coverage with perturbation-stable attention evidence.
Unlike hybrid methods~\cite{zhang2025beyond, deng2026scope} that combine saliency and diversity to improve token selection, CoRePruner assigns them distinct roles motivated by our findings: diversity provides instruction-aligned coverage, while reliability-aware attention identifies stable visual evidence.

Given visual tokens $Z=\{z_i\}_{i=1}^{n}$, CoRePruner selects a subset $S^* \subset Z$ with $|S^*| = m_1 + m_2$ through two complementary stages (Figure~\ref{fig:overview}):
\begin{equation*}
\resizebox{0.98\linewidth}{!}{
$\displaystyle
Z
\xrightarrow{\text{Task-Relevant Diversity}}
S
\xrightarrow{\text{Reliability-Aware Attention}}
S^* = S \cup A
$%
}
\end{equation*}
where $S$ contains $m_1$ task-relevant diverse tokens and $A\subset Z\setminus S$ contains $m_2$ reliability-aware attention anchors selected in Stage~II.

\subsection{Task-Relevant Diversity}

Finding~2 shows that useful evidence is not confined to a unique token subset. Stage I therefore establishes broad, instruction-aligned coverage rather than determining the final token set alone. We instantiate this role with instruction-aware determinantal point process (DPP) selection, which jointly promotes task relevance and diversity. 

Let $z_i \in \mathbb{R}^d$ denote the $i$-th visual token embedding from the vision encoder and $q \in \mathbb{R}^d$ the instruction embedding from the text encoder. We compute instruction relevance as $r_i = \operatorname{sim}(z_i, q)$ and min--max normalize it to $\tilde{r}_i \in [0,1]$. 
For CLIP-based backbones (LLaVA-1.5 and LLaVA-NeXT), we negate the similarity before normalization, i.e., $r_i=-\operatorname{sim}(z_i,q)$, because raw CLIP patch--text similarity is often higher on background patches than on the queried foreground~\cite{li2025closer, chenyang_gradeclip}. All LLaVA-1.5 and LLaVA-NeXT results use this negated form; Qwen2.5-VL results use $r_i=\operatorname{sim}(z_i,q)$ as written, with $q$ the mean of the instruction token embeddings in the language model's input space.
The relevance-weighted similarity kernel is 
\begin{equation}
    \tilde{L}_{ij} = \tilde{r}_i \cdot \kappa(z_i, z_j) \cdot \tilde{r}_j,
\end{equation}
where $\kappa(\cdot, \cdot)$ measures pairwise token similarity. We select 
\begin{equation}
    S = \arg\max_{\substack{T\subset Z\\|T|=m_1}}
    \det(\tilde L_T),
\end{equation}
where $\tilde{L}_T$ is the principal submatrix indexed by $T$. Maximizing $\det(\tilde L_T)$ discourages redundant selections, while relevance modulation favors instruction-aligned evidence. 

The formulation accommodates alternative similarity kernels, and $S$ serves as an instruction-aligned coverage set rather than a final token-importance estimate; Stage II complements it with stable attention evidence.

\subsection{Reliability-Aware Attention}

Finding~2 suggests that attention preserves complementary, perceptually salient evidence beyond the instruction-aligned coverage established by Stage~I. Guided by Finding~1, Stage~II augments the selected set with reliability-aware attention anchors that favor perturbation-stable signals.

For each remaining token $z_i\in Z\setminus S$, let $a_i$ denote its attention score on the original input. Across $K$ semantics-preserving perturbations, we compute uncertainty $u_i$ using Equation~\ref{eq:uncertainty} and define 
\begin{equation}
\rho_i=\frac{1}{1+u_i}, \qquad
\hat{a}_i=\rho_i \cdot a_i,
\end{equation}
where $a_i$ provides the saliency signal and $\rho_i\in(0,1]$ preserves more of it when attention is stable. We select the $m_2$ highest-scoring tokens outside $S$:
\begin{equation}
\label{eq: reliable anchor token selection}
    A=
    \operatorname{TopK}_{z_i\in Z\setminus S}
    \left(\hat a_i,m_2\right).
\end{equation}
The final set $S^*=S\cup A$ reflects both findings: it combines complementary evidence from different criteria and favors attention signals that remain stable under semantics-preserving variation.

\begin{table*}[t]
\centering
\footnotesize
\begin{tabular*}{\textwidth}{@{\extracolsep{\fill}} l | c c c c | c c }
\hline
\textbf{Method} & \textbf{VQA}\textsuperscript{Text} & \textbf{MME} &
\textbf{MMB}\textsuperscript{EN} & \textbf{MMB}\textsuperscript{CN} & \textbf{Avg.} & \textbf{Rel.} \\
\hline

\rowcolor{gray!20}
\multicolumn{7}{c}{\it{Upper Bound, All 1296 Tokens (100\%)}}\\
\hline
Qwen2.5-VL-7B & 84.8 & 2304 & 82.8 & 83.2 & 91.5 & 100.0\%\\
\hline

\rowcolor{gray!20}
\multicolumn{7}{c}{\it{Retain 512 Tokens \textcolor{green!60!black}{($\downarrow$ 60.5\%)}}}\\
\hline
FastV {\scriptsize\ensuremath{(\mathrm{ECCV24})}}     & \textbf{84.1} & 2317 & 82.0 & 81.1 & \underbar{90.8} & \underbar{99.1\%} \\
DivPrune {\scriptsize\ensuremath{(\mathrm{CVPR25})}}     & 81.8 & 2279 & 81.6 & \textbf{82.1} & 89.9 & 98.2\%\\
CDPruner {\scriptsize\ensuremath{(\mathrm{NeurIPS25})}}  & 82.1 & \underbar{2323} & \underbar{82.2} & 81.4 & 90.5 & 98.7\%\\ 
EADP {\scriptsize\ensuremath{(\mathrm{ECCV26})}}
& 78.6 & 2213 &  81.6 &   80.8 & 87.9 & 96.1\%\\
\midrule
\textbf{CoRePruner} {\scriptsize\ensuremath{(\mathrm{Ours})}}    & \underbar{83.6} & \textbf{2327} & \textbf{84.0} & \underbar{81.6} & \textbf{91.4} & \textbf{99.8\%} \\
\hline

\rowcolor{gray!20}
\multicolumn{7}{c}{\it{Retain 256 Tokens \textcolor{green!60!black}{($\downarrow$ 80.2\%)}}}\\
\hline
FastV {\scriptsize\ensuremath{(\mathrm{ECCV24})}}     & \textbf{81.5} & 2238 & 79.6 & 78.9 & 88.0 & 96.1\% \\
DivPrune {\scriptsize\ensuremath{(\mathrm{CVPR25})}}     & 76.0 & 2184 & 80.0 & 79.6 & 86.2 & 94.2\%\\
CDPruner {\scriptsize\ensuremath{(\mathrm{NeurIPS25})}}  & 78.0 & \underbar{2296} & \underbar{80.9} & \underbar{79.9} & \underbar{88.4} & \underbar{96.3\%} \\ 
EADP {\scriptsize\ensuremath{(\mathrm{ECCV26})}}
& 73.8 & 2202 &  80.2 &   78.5 & 85.7 & 93.5\% \\
\midrule
\textbf{CoRePruner} {\scriptsize\ensuremath{(\mathrm{Ours})}}    & \underbar{80.9} & \textbf{2332} & \textbf{83.1} & \textbf{80.6} & \textbf{90.3} & \textbf{98.5\%} \\
\hline

\rowcolor{gray!20}
\multicolumn{7}{c}{\it{Retain 128 Tokens \textcolor{green!60!black}{($\downarrow$ 90.1\%)}}}\\
\hline
FastV {\scriptsize\ensuremath{(\mathrm{ECCV24})}}     & \underbar{73.8} & 2008 & 72.9 & 72.2 & 79.8 & 87.3\%\\
DivPrune {\scriptsize\ensuremath{(\mathrm{CVPR25})}}     & 67.0 & 2108 & 77.8 & \underbar{77.8} & 82.0 & 89.5\%\\
CDPruner {\scriptsize\ensuremath{(\mathrm{NeurIPS25})}}  & 72.0 & 2127 & 76.2 & 76.5 & \underbar{82.8} & \underbar{90.3\%} \\ 
EADP {\scriptsize\ensuremath{(\mathrm{ECCV26})}}
& 65.8 & \underbar{2138} &  \underbar{78.4} &   76.4 & 81.9 & 89.2\%\\
\midrule
\textbf{CoRePruner} {\scriptsize\ensuremath{(\mathrm{Ours})}}    & \textbf{75.3} & \textbf{2290} & \textbf{80.8} & \textbf{79.0} & \textbf{87.4} & \textbf{95.2\%} \\
\hline
\end{tabular*}
\caption{
Performance comparison on Qwen2.5-VL-7B under constrained token budgets. Consistent improvements show that CoRePruner generalizes to architectures with different visual-token generation mechanisms, with margins widening as the budget tightens. Notation follows Table~\ref{tab:llava1.5}.
}
\label{tab: qwen}
\end{table*}

\newcolumntype{L}[1]{>{\raggedright\arraybackslash}m{#1}}
\newcolumntype{C}[1]{>{\centering\arraybackslash}m{#1}}
\begin{table*}[!ht]
\centering
\scriptsize
\resizebox{\textwidth}{!}{%
\begin{tabular}{L{3cm}|C{2cm}|C{2cm}*{3}{C{1.3cm}}}
\hline
\multirow{2}{*}{Method} & \textbf{MVBench} & \multicolumn{4}{c}{\textbf{VideoMME}} \\ 
& test & w/o sub & short & medium & long \\
\hline
\rowcolor{gray!20}
\multicolumn{6}{c}{\it{Upper Bound, All 6,272 Tokens (100\%)}} \\
\hline
Qwen2.5-VL-7B & 60.7 & 61.3 & 70.8 & 61.2 & 51.9 \\
\hline
\rowcolor{gray!20}
\multicolumn{6}{c}{\it{Retain 1568 Tokens \textcolor{green!60!black}{($\downarrow$ 75\%)}}} \\
\hline
VisPruner {\scriptsize\ensuremath{(\mathrm{ICCV25})}} & \textbf{59.8} & 58.7 & 68.0 & 57.6 & \textbf{50.0} \\
CDPruner {\scriptsize\ensuremath{(\mathrm{NeurIPS25})}} & 58.5 & 58.9 & 67.7 & \textbf{59.3} & 49.7 \\
\hline
\textbf{CoRePruner(Ours)} & \textbf{59.8} & \textbf{59.3} & \textbf{69.3} & 59.0 & 49.7 \\
\hline
\rowcolor{gray!20}
\multicolumn{6}{c}{\it{Retain 941 Tokens \textcolor{green!60!black}{($\downarrow$ 85\%)}}} \\
\hline
VisPruner {\scriptsize\ensuremath{(\mathrm{ICCV25})}} & 58.9 & 56.8 & 65.4 & 56.8 & 48.1 \\
CDPruner {\scriptsize\ensuremath{(\mathrm{NeurIPS25})}} & 56.1 & 57.7 & \textbf{66.9} & 57.6 & 48.6 \\
\hline
\textbf{CoRePruner(Ours)} &\textbf{59.1} & \textbf{58.2} & 66.4 & \textbf{57.8} & \textbf{51.6} \\
\hline
\rowcolor{gray!20}
\multicolumn{6}{c}{\it{Retain 627 Tokens \textcolor{green!60!black}{($\downarrow$ 90\%)}}} \\
\hline
VisPruner {\scriptsize\ensuremath{(\mathrm{ICCV25})}} & 57.0 & 56.2 & \textbf{65.1} & 55.4 & 47.9  \\
CDPruner {\scriptsize\ensuremath{(\mathrm{NeurIPS25})}} & 54.4 & 55.8 & 63.9 & \textbf{56.3} & 47.2 \\
\hline
\textbf{CoRePruner (Ours)} & \textbf{58.1} & \textbf{57.0} & \textbf{65.1} & 56.2 & \textbf{49.8} \\
\hline
\end{tabular}%
}
\caption{
Video-understanding performance on MVBench and VideoMME with Qwen2.5-VL-7B.
CoRePruner matches or exceeds the strongest baseline on most metrics, with clearer gains under 85\% and 90\% token reduction.
}
\label{tab:video}
\end{table*}

\section{Experiments}
We test whether reliable token-set construction improves pruning across tasks, token budgets, and architectures.
Implementation details, including token allocation and perturbation configurations, are provided in Appendices~\ref{sec:app-token-allocation}--\ref{sec:app-software-evaluation}.

\paragraph{Setup.}
To assess generalizability across different visual-token generation mechanisms, we evaluate CoRePruner on representative MLLM architectures. LLaVA-1.5~\cite{liu2024improved} uses a CLIP vision encoder and a Vicuna-based language model~\cite{chiang2023vicuna}. 
LLaVA-NeXT~\cite{liu2024llavanext} extends LLaVA-1.5 with dynamic high-resolution slicing. 
Qwen2.5-VL~\cite{Qwen2.5-VL} employs native dynamic-resolution encoding that produces variable-length visual tokens.

Baselines include FastV~\cite{chen2024image}, SparseVLM~\cite{zhang2024sparsevlm}, PDrop~\cite{xing2024pyramiddrop}, VisionZip~\cite{yang2025visionzip}, DivPrune~\cite{alvar2025divprune}, VisPruner~\cite{zhang2025beyond}, CDPruner~\cite{zhang2026beyond}, and EADP~\cite{wang2026combating}.
Together, they span attention-based, diversity-based, and hybrid selection strategies.

\paragraph{Benchmarks and Metrics.}
The evaluation spans eleven image- and video-understanding benchmark settings: VQAv2~\cite{goyal2017making}, GQA~\cite{hudson2019gqa}, TextVQA~\cite{singh2019towards}, MME~\cite{fu2026mme}, MMBench-EN/CN~\cite{liu2024mmbench}, MMVet~\cite{yu2023mm}, ScienceQA-IMG~\cite{lu2022learn}, POPE~\cite{li2023evaluating}, MVBench~\cite{li2024mvbench}, and VideoMME~\cite{fu2025video}.
Accuracy (\%) is reported for VQAv2, GQA, TextVQA, MMBench-EN/CN, ScienceQA-IMG, and MVBench. MME uses the Perception (P) score (Total P+C for Qwen2.5-VL), MMVet uses an LLM-judge score, and POPE uses F1 score. For VideoMME, results include the overall score without subtitles and its short-, medium-, and long-video breakdowns.

\begin{figure}[!t]
    \centering
    \includegraphics[width=1.0\linewidth]{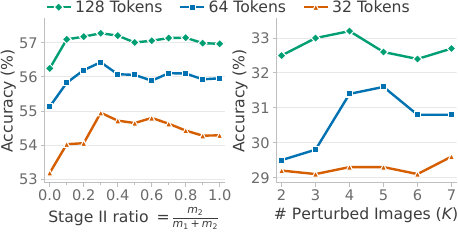}
    \caption{
        Effects of token allocation and perturbation count.
        \textbf{Left:} Accuracy generally peaks near a Stage~II ratio of $0.3$ across token budgets.
        \textbf{Right:} Performance versus the number of perturbed images
        $K$ used to estimate attention reliability, with fluctuations when $K\ge 4$.
    }
    \label{fig:ablation_accuracy}
\end{figure}

\begin{figure}[!t]
    \centering
    \includegraphics[width=1.0\linewidth]{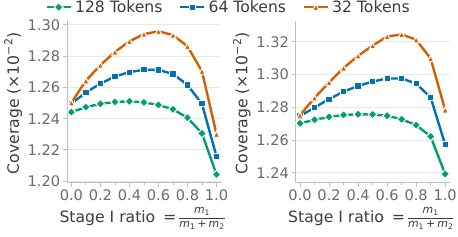}
    \caption{
        Feature-space coverage of retained tokens, measured by the RMS-std of their features, across Stage~I ratios on GQA (Left) and TextVQA (Right) with LLaVA-1.5-7B. Coverage peaks at intermediate ratios, supporting complementary token selection.
    }
    \label{fig:div-cov}
\end{figure}

\subsection{Main Results}


We first evaluate CoRePruner on LLaVA-1.5-7B, a widely adopted backbone for visual token pruning, under strict token budgets retaining 128, 64, or 32 visual tokens (Table~\ref{tab:llava1.5}).

CoRePruner variants attain the best average score at every budget, and their margin over the strongest baseline is generally larger under tighter budgets: $+0.7$ performance-retention points at 128 tokens, versus $+1.7$ and $+1.1$ at 64 and 32. 
This pattern is consistent with Finding~1, where the advantage of removing high-uncertainty tokens over low-attention pruning widens under tighter budgets.
These results suggest that reliability becomes more important as fewer tokens
survive.

To check that the gains do not depend on a particular diversity formulation, we instantiate Stage~I with cosine, RBF, and polynomial kernels. The three variants differ by at most $0.2$ average points at every budget, showing that CoRePruner is not tied to a specific similarity measure.

On Qwen2.5-VL-7B and LLaVA-NeXT-7B, as shown in Table~\ref{tab: qwen} and Appendix~\ref{sec:llava-next-results} (Table~\ref{tab: llava-next}), CoRePruner improves overall performance under matched token budgets, indicating that reliable token-set construction transfers across architectures with different visual-token generation mechanisms. These gains also extend to video benchmarks (Table~\ref{tab:video}), where CoRePruner matches or outperforms the strongest baseline on most MVBench and VideoMME metrics, particularly under 85\% and 90\% token reduction, showing that the construction is not limited to single-image inputs and remains effective with longer visual sequences.

\begin{table}[!t]
\centering
\small
\setlength{\tabcolsep}{2pt} 
\begin{tabular*}{\columnwidth}{@{\extracolsep{\fill}} l | c c c c}
\toprule
\multirow{2}{*}{\textbf{Method}}
& \textbf{Prefill}
& \textbf{Decode}
& \textbf{KV}
& \textbf{GPU} \\[-2pt]
& \textbf{Time}
& \textbf{Time}
& \textbf{Cache}
& \textbf{Mem.} \\
\midrule
LLaVA-NeXT-7B & 254.7 & 31.0 & 1465.3 & 17.1\\
\midrule
VisPruner  & 35.4  & 20.0 & 185.3 & 15.2\\
CDPruner  & 35.4 & 20.0 & 185.3 & 14.4\\
\midrule
\textbf{CoRePruner} & 35.7 & 20.0 & 185.3 & 16.6\\
\bottomrule
\end{tabular*}
\caption{Efficiency analysis on LLaVA-NeXT-7B with 320 tokens. CoRePruner matches prior pruning methods in prefill and decode latency and KV cache size, with additional GPU memory overhead.  Prefill is reported per input and decode per generated token (ms); KV cache in MB, and memory in GB. Prefill and decode exclude vision encoding and token selection.}
\label{tab: efficiency}
\end{table}

\begin{table}[!t]
\centering
\small
\setlength{\tabcolsep}{2pt}
\begin{tabularx}{\columnwidth}{
    l |
    >{\centering\arraybackslash}X
    >{\centering\arraybackslash}X
    >{\centering\arraybackslash}X |
    >{\centering\arraybackslash}X
}
\toprule

\multirow{2}{*}{\textbf{Benchmark}}
& \textbf{Stage I}
& \textbf{Stage II}
& \textbf{w/o}
& \multirow{2}{*}{\textbf{Ours}} \\

& \textbf{-only}
& \textbf{-only}
& \textbf{Reliability}
& \\



\hline
\rowcolor{gray!20}
\multicolumn{5}{c}{
    \itshape
    Retain 64 Tokens
    \textcolor{green!60!black}{($\downarrow$ 88.9\%)}
} \\
\hline

{VQA}\textsuperscript{Text}
    & 55.1 & 55.9 & 56.1 & 56.4 \\
GQA
    & 58.7 & 56.0 & 58.6 & 59.0 \\
MME
    & 1410.2 & 1437.6 & 1430.0 & 1438.2 \\
MMB\textsuperscript{EN}
    & 61.2 & 61.3 & 61.3 & 61.4 \\
MMVet
    & 27.9 & 30.5 & 30.4 & 31.4 \\

\hline
\rowcolor{gray!20}
\multicolumn{5}{c}{
    \itshape
    Retain 32 Tokens
    \textcolor{green!60!black}{($\downarrow$ 94.4\%)}
} \\
\hline

{VQA}\textsuperscript{Text}
    & 53.2 & 54.2 & 54.7 & 55.0 \\
GQA
    & 57.1 & 52.3 & 56.8 & 57.1 \\
MME
    & 1371.8 & 1307.0 & 1358.5 & 1380.6 \\
MMB\textsuperscript{EN}
    & 59.5 & 59.5 & 59.8 & 60.2 \\
MMVet
    & 27.8 & 25.7 & 27.4 & 29.3 \\
\bottomrule
\end{tabularx}
\caption{
Ablation analysis on LLaVA-1.5-7B.
\textbf{Stage I-only} and \textbf{Stage II-only} use a single stage; \textbf{w/o Reliability} replaces the Stage II score with raw attention. Both stages contribute, and removing the stability weighting lowers accuracy on every metric at both budgets.
}
\label{tab:ablation-study}
\end{table}

\section{Analysis}
\label{sec:analysis}

\paragraph{Effect of Token Allocation between Stage I and Stage II.}
We analyze the effect of Stage-II token allocation on TextVQA by varying the ratio $m_2/(m_1+m_2)$ while keeping the total number of retained tokens fixed.
As shown in Figure~\ref{fig:ablation_accuracy} (left), performance consistently peaks when about $30\%$ of tokens are allocated to Stage II, whereas allocating more tokens degrades performance due to reduced instruction-aware contextual coverage. This result suggests that a moderate number of reliability-aware tokens effectively complements instruction-aware diversity-based token selection. Detailed implementation settings for token allocation are provided in Appendix~\ref{sec:app-token-allocation}.

\paragraph{Effect of the Number of Perturbed Images.}
We analyze the effect of the number of perturbed images $K$ used to estimate attention reliability on MMVet. 
Figure~\ref{fig:ablation_accuracy} (right) shows that performance generally improves up to $K=4$, suggesting that combining multiple perturbation types yields a more stable reliability estimate. Beyond $K=4$, performance fluctuates, indicating diminishing returns despite the computational cost.
Therefore, we use $K=4$ in all experiments, showing that reliability estimation requires only a small number of perturbations. Detailed perturbation types and intensities are provided in Appendix~\ref{sec:app-perturbations}.

\paragraph{Efficiency Analysis.}
We evaluate inference efficiency on LLaVA-NeXT-7B using the POPE benchmark, following prior work~\cite{zhang2026beyond}, on a single NVIDIA A100-80GB GPU. Table~\ref{tab: efficiency} reports prefill time, decode time, KV cache size, and GPU memory consumption under a 320-token budget. 
CoRePruner matches existing pruning methods in prefill time, decode time, and KV cache size, as these depend only on the number of retained tokens.
Including vision encoding and token selection, CoRePruner reduces end-to-end latency by 27.9\% (322.8\,ms to 232.8\,ms) and FLOPs by 64.2\% (42.64T to 15.26T) relative to the unpruned model.
This cost includes the $K{=}4$ perturbed vision-encoder passes used for reliability estimation ($4\times2.56$T FLOPs), which VisPruner and CDPruner do not require (109.0\,ms and 111.3\,ms; 5.02T FLOPs).
The comparisons in Tables~\ref{tab:llava1.5}--\ref{tab: qwen} are therefore made at equal token count rather than equal total compute.
Together with the performance gains in these tables, CoRePruner offers a favorable accuracy--efficiency trade-off relative to the unpruned model, at the cost of additional vision-encoder computation and memory compared with prior pruning methods.

\paragraph{Stage-wise Ablation Study.}
To evaluate the complementary roles of each stage, we conduct a stage-wise ablation study. Table~\ref{tab:ablation-study} compares Stage I only, Stage II only, and Stage I+II on LLaVA-1.5-7B across multiple benchmarks. Stage I+II achieves the best overall performance, consistently outperforming either stage alone. This result supports our observations that reliable token selection benefits from combining instruction-aligned coverage (Stage I) with stable visual evidence captured by reliability-aware attention (Stage II). Together, they form a more reliable token set for robust pruning.

\paragraph{Reliability-aware Attention Ablation Study.}
To examine the role of reliability in Stage II, we compare attention-based selection with and without reliability-aware reweighting. Table~\ref{tab:ablation-study} reports results on LLaVA-1.5-7B across multiple benchmarks. Across all token budgets, incorporating reliability consistently improves performance, with gains becoming more pronounced under aggressive pruning. These results suggest that attention magnitude alone is insufficient for reliable token selection. By suppressing unstable tokens and emphasizing attention consistency under minor perturbations, reliability-aware selection yields more informative token subsets, particularly in low token budgets.

\paragraph{Semantic Coverage.}
To analyze how complementary token selection affects semantic coverage, we use the Root-Mean-Square standard deviation (RMS-std) of visual-token features, where higher RMS-std indicates broader coverage. We vary the Stage~I ratio $m_1/(m_1+m_2)$, with the remaining tokens selected by Stage~II.
Figure~\ref{fig:div-cov} reports results on GQA and TextVQA across token budgets. Coverage typically peaks at an intermediate ratio and decreases as the Stage~I ratio approaches $1.0$.
This pattern suggests that neither diversity-only nor attention-only selection maximizes feature-space coverage, supporting their complementary use. 
\section{Conclusion}

We revisit visual token pruning through the lens of selection reliability. Our analysis reveals two limitations of existing criteria: attention-based importance is locally unstable under semantics-preserving perturbations, while different pruning criteria are globally inconsistent, selecting largely disjoint token subsets despite comparable downstream performance. Together, these findings show that useful visual evidence is neither determined by attention magnitude alone nor confined to a unique token set.
Guided by these findings, we introduce CoRePruner, a training-free framework that constructs reliable token sets by assigning complementary roles to instruction-aware diversity and reliability-aware attention. Across multiple vision-language benchmarks, CoRePruner consistently improves overall performance, with larger gains under aggressive pruning. These results highlight reliability, beyond token importance alone, as a central principle for designing and evaluating visual token pruning methods.

\section*{Limitations}
CoRePruner requires multiple perturbed inputs to estimate attention stability, which increases vision-encoder computation and peak GPU memory; this cost is fixed per image and independent of the token budget, so its relative share grows as the budget tightens.
As a result, at the same token count, CoRePruner has higher end-to-end latency than pruning methods that encode only the clean image (Section~\ref{sec:analysis}).
Although the perturbations are designed to preserve task-relevant semantics, this assumption may not hold for fine-grained recognition or text-centric tasks.
The proposed uncertainty measure may also be sensitive to tokens with small mean attention, since it normalizes by the mean.
Our video evaluation is limited to Qwen2.5-VL-7B on two benchmarks, so conclusions about video understanding should be drawn with caution.
Future work should investigate more efficient stability estimation and broader notions of token-selection reliability.

\section*{Ethical Considerations}

This work uses publicly available benchmarks and pretrained models and does not collect new human-subject data or annotations. Visual token pruning can reduce the computational cost of processing long visual sequences, and we report the efficiency trade-offs introduced by reliability estimation.
Pruning necessarily removes part of the visual input, and under aggressive budgets, task-relevant evidence may be omitted. Our findings further show that the retained evidence depends on the selection criterion, so pruned models may operate on incomplete visual context. Improving token-selection reliability does not eliminate biases, hallucinations, or harmful outputs inherited from
the underlying MLLMs and evaluation datasets. The proposed method should therefore not be interpreted as providing safety guarantees in high-stakes applications.
\section*{Acknowledgments}
This work was supported by the Institute of Information and Communications Technology Planning and Evaluation (IITP) grant funded by the Korean government (MSIT) (No. RS-2025-02263031, Development of On-Device AI Cooperative Actions (Perception, Decision-Making, and Response) Among Networked Devices).

\clearpage

\bibliography{custom}

\appendix

\clearpage
\twocolumn
\startcontents[appendix]

\begingroup

\hypersetup{
    colorlinks=true,
    linkcolor=blue,
    linktoc=all
}

\begin{center}
    {\LARGE\bfseries APPENDIX}
\end{center}

\vspace{0.8em}
\linespread{1.08}\selectfont

\titlecontents{section}
    [0em]
    {\small\bfseries\addvspace{0.75em}}
    {\contentslabel{2.3em}}
    {}
    {\titlerule*[0.6pc]{.}\contentspage}

\titlecontents{subsection}
    [2.5em]
    {\small\addvspace{0.22em}}
    {\contentslabel{3.2em}}
    {}
    {\titlerule*[0.6pc]{.}\contentspage}

\titlecontents{subsubsection}
    [5.7em]
    {\small\addvspace{0.12em}}
    {\contentslabel{4.3em}}
    {}
    {} 

\printcontents[appendix]{}{1}{
    \setcounter{tocdepth}{3}
}
\endgroup

\newpage


\newcolumntype{C}[1]{>{\centering\arraybackslash}m{#1}}

\section{Implementation Details}
\subsection{Findings}
All analyses in this section are conducted using LLaVA-1.5-7B~\cite{liu2024improved}. The attention visualizations in Figure~\ref{fig:attention-vis} of the main manuscript are obtained on VQAv2~\cite{goyal2017making}, where attention maps are averaged across all heads and extracted from the last four layers of the vision encoder. The analyses in Figure~\ref{fig:pruning-uncertainty}(a) and Figure~\ref{fig:pruning-uncertainty}(b) are performed on the GQA dataset~\cite{hudson2019gqa}. All experiments are conducted on a single NVIDIA A100 GPU.


\subsubsection{Finding 1: Attention-Based Importance Is Locally Unstable}
\label{sec:app-observation1}
To ensure a fair comparison between tokens with different uncertainty levels, the high-uncertainty and low-uncertainty token groups used in Figure~\ref{fig:pruning-uncertainty}(a) are constructed to have closely matched attention-score statistics. In particular, we match the mean attention magnitude of the two groups so that any performance difference cannot be attributed to stronger raw attention. As shown in Table~\ref{tab:attn_distribution}, the two groups exhibit comparable mean attention scores and standard deviations across pruning budgets, while the high-uncertainty group consistently shows larger uncertainty values than the low-uncertainty group. This controlled setup isolates the effect of attention uncertainty and reduces the possibility that the performance differences observed in Figure~\ref{fig:pruning-uncertainty}(a) are simply due to differences in attention magnitude. 
Figure~\ref{fig:pruning-uncertainty}(b) provides a complementary analysis by fixing each pruning budget and comparing the removal of low-, middle-, and high-uncertainty tokens under the same uncertainty definition.

\subsubsection{Finding 2: Pruning Criteria Are Globally Inconsistent}
\label{sec:app-observation2}
To ensure a fair comparison between different pruning criteria, CLS-token-based attention and instruction-aware diversity are compared on the same image-question pairs at each pruning budget in Table~\ref{tab:attention_vs_diversity} and Figure~\ref{fig:iou_visualize} of main manuscript. For each sample, we separately apply CLS-token-based attention and instruction-aware diversity, and then measure the overlap between resulting selected token sets using Intersection-over-Union (IoU). The qualitative examples in Figure~\ref{fig:iou_visualize} correspond to a retained token budget of 128 and are drawn from VQAv2, GQA, and TextVQA~\cite{singh2019towards}.

\begin{table}[!t]
\centering
\small
\setlength{\tabcolsep}{3pt}
\begin{tabular*}{\columnwidth}{@{\extracolsep{\fill}} c | c | c c | c c}
\toprule
\multirow{2}{*}{\textbf{\# Tokens}}
& \multirow{2}{*}{\textbf{Groups}}
& \multicolumn{2}{c|}{\textbf{Attention Score}}
& \multicolumn{2}{c}{\textbf{Uncertainty}} \\ 
&
& Mean
& Std.
& Mean
& Std. \\
\midrule
\multirow{2}{*}{128}
& $U\uparrow$   & 0.0029 & 0.0105 & 0.0028 & 0.0120 \\
& $U\downarrow$ & 0.0024 & 0.0072 & 0.0010 & 0.0067 \\
\midrule
\multirow{2}{*}{256}
& $U\uparrow$   & 0.0014 & 0.0068 & 0.0018 & 0.0106 \\
& $U\downarrow$ & 0.0012 & 0.0047 & 0.0005 & 0.0044 \\
\midrule
\multirow{2}{*}{320}
& $U\uparrow$   & 0.0011 & 0.0058 & 0.0014 & 0.0091 \\
& $U\downarrow$ & 0.0010 & 0.0040 & 0.0009 & 0.0073 \\
\midrule
\multirow{2}{*}{448}
& $U\uparrow$   & 0.0011 & 0.0058 & 0.0014 & 0.0091 \\
& $U\downarrow$ & 0.0010 & 0.0040 & 0.0009 & 0.0073 \\
\midrule
\multirow{2}{*}{512}
& $U\uparrow$   & 0.0011 & 0.0054 & 0.0013 & 0.0085 \\
& $U\downarrow$ & 0.0010 & 0.0038 & 0.0010 & 0.0072 \\
\bottomrule
\end{tabular*}

\caption{Attention and uncertainty statistics of the high-uncertainty ($U\uparrow$) and low-uncertainty ($U\downarrow$) token groups used in Figure~\ref{fig:pruning-uncertainty}(a). For a fair comparison, the two groups are constructed to have closely matched attention-score distributions, including similar mean attention magnitudes, allowing the effect of uncertainty to be examined while controlling for raw attention strength.}
\label{tab:attn_distribution}
\end{table}

\subsection{Experiments}
\subsubsection{Token Allocation.}
\label{sec:app-token-allocation}
\begin{figure}[t]
    \centering
    \includegraphics[width=0.9\linewidth]{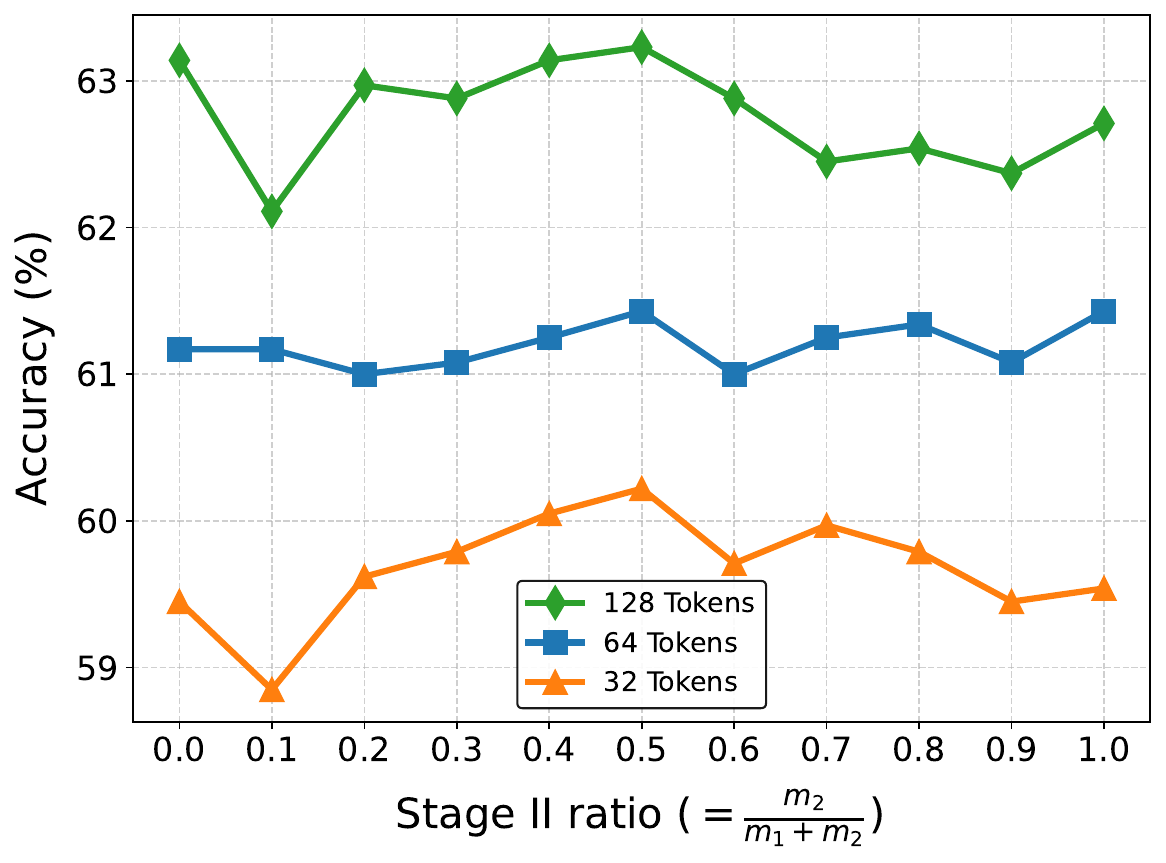}
    \caption{
    Ablation of the Stage-II token allocation ratio on MMBench.
    Performance peaks at 0.5, indicating that reasoning-centric benchmarks benefit from a larger allocation to reliability-aware tokens.
    }
    \label{fig:supp_ratio_mmbench}
\end{figure}

As discussed in Analysis section, the optimal balance between Stage-I ($m_1$) and Stage-II ($m_2$) token selection is determined by the specific demands of each task. We categorize the benchmarks into two distinct groups and determine the allocation ratio, $m_2/(m_1+m_2)$, based on representative experiments for each category.
The first group includes benchmarks that primarily require concise, evidence-based responses, typically guided by the instruction: \emph{``Answer the question using a single word or phrase.''}. This category encompasses VQAv2, TextVQA, GQA, POPE~\cite{li2023evaluating}, and MME~\cite{fu2026mme}. Using TextVQA as a representative case for this group, we analyzed the effect of the allocation ratio in Figure~\ref{fig:ablation_accuracy}. The results show that performance peaks at a Stage-II ratio of approximately 30\% ($m_2/(m_1+m_2) = 0.3$). Based on this observation, we applied the 0.3 ratio to all benchmarks in this category, as they share the common requirement of capturing precise, localized instruction-aware evidence (Stage I) for short-form answering.
The second group consists of benchmarks that involve complex reasoning or multiple-choice formats, including MMBench-EN/CN~\cite{liu2024mmbench}, MMVet~\cite{yu2023mm}, and ScienceQA~\cite{lu2022learn}. For this category, we conducted a representative ablation study on MMBench, as illustrated in Figure~\ref{fig:supp_ratio_mmbench}. We found that a higher Stage II allocation ratio of 50\% ($m_2/(m_1+m_2) = 0.5$) is more effective, indicating that these tasks benefit from a greater emphasis on reliability-aware tokens to preserve global context and salient object features. Accordingly, this 0.5 ratio was applied to the remaining reasoning-centric benchmarks.
In both groups, the ratio is chosen on a single representative benchmark and applied unchanged to the other benchmarks in the group. To verify that the results do not depend on this grouping, we also apply a single ratio of 0.3 to all benchmarks: at 64 tokens on LLaVA-1.5-7B, the average score is 62.9, compared with 63.2 under the grouped allocation and 62.3 for the strongest baseline.

\begin{table}[!t]
\centering
\scriptsize
\setlength{\tabcolsep}{4.0pt} 
\renewcommand{\arraystretch}{1.08}
\begin{tabular}{@{}lll@{}}
\toprule
\textbf{Package} & \textbf{Version} & \textbf{Primary use} \\
\midrule
Python          & 3.10.19 & Runtime environment \\
PyTorch         & 2.1.2   & Model inference \\
torchvision     & 0.16.2  & Image transformations \\
Transformers    & 4.37.2  & Model loading and generation \\
NumPy           & 1.22.3  & Perturbation and numerical operations \\
Pillow          & 10.4.0  & Image preprocessing \\
qwen-vl-utils   & 0.0.14  & Qwen input processing \\
\bottomrule
\end{tabular}
\caption{Core software packages used for model inference and evaluation.}
\label{tab:software-environment}
\end{table}

\subsubsection{Perturbations.}
\label{sec:app-perturbations}
To evaluate token reliability under input variations, we generate perturbed inputs using a set of mild, semantics-preserving transformations, including resolution degradation, color filtering, Gaussian blur, additive noise, brightness and saturation adjustments, and horizontal flipping. All perturbations are applied with mild magnitudes to preserve the semantic content of the image while introducing small appearance-level variations. Specifically, we use down-up sampling with a scale factor of 0.75, blue-tint blending with coefficient 0.18, Gaussian blur with radius 1.5, additive Gaussian noise with standard deviation $\sigma=10$, brightness adjustment with factor 1.25, saturation adjustment with factor 1.55, and horizontal flipping.

All main experiments use $K=4$, consisting of the first four perturbations in the order listed above: down-up sampling, blue-tint blending, Gaussian blur, and additive Gaussian noise. Brightness, saturation, and horizontal flipping are used only when $K>4$ (Figure~\ref{fig:ablation_accuracy}, right). The perturbation set, order, and magnitudes are fixed across all images, benchmarks, and models. These perturbations preserve task-relevant semantics: without pruning, LLaVA-1.5-7B obtains 61.9 on GQA with clean inputs and 62.0--62.3 with perturbed inputs, and prediction correctness is unchanged on 94.4--95.7\% of questions.

\begin{table}[!t]
\centering
\scriptsize
\setlength{\tabcolsep}{3.0pt}
\renewcommand{\arraystretch}{1.09}
\begin{tabularx}{\columnwidth}{
@{}
>{\raggedright\arraybackslash}p{1.75cm}
>{\raggedright\arraybackslash}p{2.35cm}
>{\raggedright\arraybackslash}X
@{}
}
\toprule
\textbf{Artifact}
& \textbf{Checkpoint / Scope}
& \textbf{License / Usage terms} \\
\midrule
\rowcolor{gray!20}
\multicolumn{3}{@{}l}{\textit{Model checkpoints}} \\
LLaVA-1.5-7B
& \path{liuhaotian/llava-v1.5-7b}
& Llama 2 Community License Agreement \\
LLaVA-NeXT-7B
& \path{liuhaotian/llava-v1.6-vicuna-7b}
& Llama 2 Community License Agreement \\
Qwen2.5-VL-7B
& \path{Qwen/Qwen2.5-VL-7B-Instruct}
& Apache License 2.0 \\

\midrule
\rowcolor{gray!20}
\multicolumn{3}{@{}l}{\textit{Image-understanding benchmarks}} \\
VQAv2
& Annotations and COCO images
& CC BY 4.0 for VQA annotations; COCO image licenses apply \\
GQA
& Questions, scene graphs, and source images
& No standalone license stated on the official release;
source-image terms apply \\
TextVQA
& Data and Open Images content
& CC BY 4.0 for TextVQA data; Open Images licenses apply \\
MME
& Benchmark data and source images
& Apache License 2.0 for the official release;
underlying image rights may vary \\
MMBench-EN/CN
& Repository and benchmark content
& Apache License 2.0 for the official repository;
third-party source-content terms apply \\
MMVet
& Evaluator and dataset
& Apache License 2.0 for code; CC BY-NC 4.0 for data \\
ScienceQA
& Code and dataset
& MIT License for code; CC BY-NC-SA 4.0 for data \\
POPE
& Repository, annotations, and COCO images
& MIT License for the repository and released annotations;
COCO image licenses apply \\
\midrule
\rowcolor{gray!20}
\multicolumn{3}{@{}l}{\textit{Video-understanding benchmarks}} \\
MVBench
& Benchmark release and source videos
& MIT License for the MVBench release;
source-video dataset licenses apply \\
VideoMME
& Dataset and source videos
& Academic research only; commercial use is prohibited;
redistribution and modification require prior approval \\
\bottomrule
\end{tabularx}
\caption{
Model checkpoints and applicable licenses and usage terms for the models
and evaluation benchmarks used in our experiments.
}
\label{tab:model-settings}
\end{table}

\subsubsection{Software and Evaluation Settings.}
\label{sec:app-software-evaluation}

The experimental setup is summarized in two tables.
Table~\ref{tab:software-environment} lists the core software packages
and their versions, while Table~\ref{tab:model-settings} provides the model
checkpoints together with the applicable licenses and terms of use for the
models and evaluation benchmarks.
Across all experiments, we use the official checkpoint-specific processors
for input preprocessing and greedy decoding for generation.
Performance is then computed using official or author-released benchmark
evaluators, including the public MM-Vet evaluator.
For the efficiency experiments reported in Table~\ref{tab: efficiency}, we additionally use
FlashAttention 2.6.3.

For datasets without a standalone license, we follow the official access
terms and the applicable licensing terms for the underlying images or videos.
Details of the evaluation benchmark splits and the corresponding
numbers of evaluation samples are provided in
Table~\ref{tab:dataset_statistics}.

\begin{table}[!t]
\centering
\small
\setlength{\tabcolsep}{4pt}
\begin{tabular*}{\columnwidth}{@{\extracolsep{\fill}} l c c}
\toprule
\textbf{Benchmark} & \textbf{Evaluation Split} & \textbf{\# Eval Samples} \\
\midrule
VQAv2      & test-dev2015        & 107,394 \\ 
GQA        & test-dev (balanced) & 12,578 \\ 
TextVQA    & val                 & 5,734 \\ 
MME        & test                & 2,374 \\ 
MMBench-EN & dev                 & 1,164 \\ 
MMBench-CN & dev                 & 1,164 \\ 
MMVet      & test                & 218 \\ 
ScienceQA  & test                & 4,241 \\ 
POPE       & COCO val2014 test   & 9,000 \\
MVBench    & test                & 4,000\\ 
VideoMME   & test                & 2,700\\ 
\bottomrule
\end{tabular*}
\caption{Evaluation benchmarks, evaluation splits, and the number of evaluation samples used in our experiments.}
\label{tab:dataset_statistics}
\end{table}

\begin{algorithm}[t]
    \caption{CoRePruner}
    \label{alg:corepruner}
    \fontsize{9.8pt}{11pt}\selectfont       
    {\raggedright
    \textbf{Input}: visual tokens $Z=\{z_i\}_{i=1}^{n}$, instruction embedding $q$,
    $K$ semantics-preserving perturbations\par
    \textbf{Parameters}: diversity budget $m_1$, anchor budget $m_2$\par
    \textbf{Output}: reliable token set $S^*$, $|S^*|=m_1+m_2$\par\smallskip
    }
    \begin{algorithmic}[1]
    \item[] \rule[-0.5em]{0pt}{1.5em}
    \textit{Stage I: Task-relevant Diversity}
    \STATE $r_i \gets \mathrm{sim}(z_i,q)$,
    normalized to $\tilde{r}_i\in[0,1]$
    \STATE $\tilde{L}_{ij}
    \gets \tilde{r}_i\,\kappa(z_i,z_j)\,\tilde{r}_j$
    \STATE $S \gets \arg\max_{|T|=m_1}\det(\tilde{L}_T)$
    \item[] \rule[-0.5em]{0pt}{1.5em}
    \textit{Stage II: Reliability-aware Attention}
    \STATE Obtain $\{a_i^{k}\}_{k=1}^{K}$ from the $K$ perturbations
    \STATE $u_i \gets$ attention uncertainty
    (Eq.~\ref{eq:uncertainty})
    \STATE $\rho_i \gets (1+u_i)^{-1}, \quad \hat{a}_i \gets \rho_i a_i$
    \STATE $A \gets \operatorname{TopK}_{z_i\in Z\setminus S}(\hat{a}_i)$, \quad $|A|=m_2$
    \STATE \textbf{return} $S^* \gets S\cup A$
    \end{algorithmic}
\end{algorithm}

\begin{table*}[!ht]
\centering
\footnotesize
\begin{tabular*}{\textwidth}{@{\extracolsep{\fill}} l | c c c c c c | c c}
\hline
\textbf{Method} & \textbf{VQA}\textsuperscript{v2} & \textbf{GQA} & \textbf{VQA}\textsuperscript{Text} & \textbf{MME} & \textbf{SQA}\textsuperscript{Img} & \textbf{POPE} & \textbf{Avg.} & \textbf{Rel.} \\
\hline
\rowcolor{gray!20}
\multicolumn{9}{c}{\it{Upper Bound, All 2880 Tokens (100\%)}}\\
\hline
LLaVA-NeXT-7B & 81.3 & 62.5 & 60.3 & 1511.8 & 67.5 & 86.8 & 72.3 & 100\% \\
\hline
\rowcolor{gray!20}
\multicolumn{9}{c}{\it{Retain 640 Tokens \textcolor{green!60!black}{($\downarrow$ 77.8\%)}}}\\
\hline
FastV {\scriptsize\ensuremath{(\mathrm{ECCV24})}}        & 77.0 & 58.9 & 58.1 & 1412.6 & 67.4 & 79.5 & 68.6 & 95.0\% \\
SparseVLM {\scriptsize\ensuremath{(\mathrm{ICML25})}}     & 77.1 & 60.3 & 57.8 & 1456.8 & 67.6 & 85.3  & 70.2 & 97.0\% \\
PDrop {\scriptsize\ensuremath{(\mathrm{CVPR25})}}        & 79.1 & 60.0 & 57.8 & 1475.9 & 66.7 & 83.8  & 70.2 & 97.0\% \\
VisionZip {\scriptsize\ensuremath{(\mathrm{CVPR25})}}    & 79.1 & 61.3 & 60.2 & 1493.4 & \underbar{68.1} & 86.0  & 71.6 & 99.0\% \\
DivPrune {\scriptsize\ensuremath{(\mathrm{CVPR25})}}     & 79.3 & 61.9 & 57.0 & 1469.7 & 67.8 & 86.9  & 71.1 & 98.1\% \\
CDPruner {\scriptsize\ensuremath{(\mathrm{NeurIPS25})}}  & 79.9 & \underbar{62.6} & 58.4 & 1474.5 & 67.9 & \underbar{87.3} & 71.6 & 99.0\% \\ 
EADP {\scriptsize\ensuremath{(\mathrm{ECCV26})}}
& \textbf{80.6} & \textbf{62.7} & 59.2 &  1494.7
& \underbar{68.0} & \textbf{87.4} & 72.1 & 99.7\%\\ \midrule
\textbf{CoRePruner} {\scriptsize\ensuremath{(\mathrm{Ours})}}    & 80.0 & 62.3 & 60.6 & 1499.5 & \textbf{68.5} & 86.9 & \underbar{72.2} & \underbar{99.9\%} \\
\textbf{CoRePruner$^{\dagger}$} {\scriptsize\ensuremath{(\mathrm{Ours})}}    & \underbar{80.1} & 62.4 & \underbar{60.7} & \underbar{1500.8} & \textbf{68.5} & 86.9  & \textbf{72.3} & \textbf{100.0\%}  \\
\textbf{CoRePruner$^{\ddagger}$} {\scriptsize\ensuremath{(\mathrm{Ours})}}    & \underbar{80.1} & 62.4 & \textbf{60.8} & \textbf{1503.6} & \textbf{68.5} & 87.1 & \textbf{72.3} & \textbf{100.0\%}  \\
\hline
\rowcolor{gray!20}
\multicolumn{9}{c}{\it{Retain 320 Tokens \textcolor{green!60!black}{($\downarrow$ 88.9\%)}}}\\
\hline
FastV {\scriptsize\ensuremath{(\mathrm{ECCV24})}}        & 61.5 & 49.8 & 52.2 & 1099.0 & 66.6 & 49.5 & 55.8 & 78.4\% \\
SparseVLM {\scriptsize\ensuremath{(\mathrm{ICML25})}}     & 73.4 & 57.7 & 55.9 & 1386.1 & 67.3 & 76.9 & 66.8 & 92.5\% \\
PDrop {\scriptsize\ensuremath{(\mathrm{CVPR25})}}        & 66.8 & 50.4 & 49.0 & 1171.5 & 66.7 & 60.8 & 58.7 & 81.7\% \\
DivPrune {\scriptsize\ensuremath{(\mathrm{CVPR25})}}     & 77.2 & 61.1 & 56.2 & 1423.3 & 67.7 & 84.7 & 69.7 & 96.3\% \\
CDPruner {\scriptsize\ensuremath{(\mathrm{NeurIPS25})}}  & 78.4 & \underbar{61.6} & \underbar{57.4} & 1453.0 & \textbf{67.8} & \textbf{87.2} & 70.8 & 97.9\% \\ 
EADP {\scriptsize\ensuremath{(\mathrm{ECCV26})}}
& \underbar{78.6} & \textbf{62.2} & 57.0 &  \underbar{1491.3}
& 67.6 & 86.9 & \underbar{71.1} & \underbar{98.3}\% \\ \midrule
\textbf{CoRePruner} {\scriptsize\ensuremath{(\mathrm{Ours})}}    & \underbar{78.6} & 61.4 & \textbf{59.5} & 1491.1 & \underbar{67.7} & 87.0 & \textbf{71.5} & \textbf{98.9\%} \\
\textbf{CoRePruner$^{\dagger}$} {\scriptsize\ensuremath{(\mathrm{Ours})}}    & \textbf{78.7} & 61.5 & \textbf{59.5} & \textbf{1501.3} & 67.4 & 87.0 & \textbf{71.5} & \textbf{98.9\%} \\
\textbf{CoRePruner$^{\ddagger}$} {\scriptsize\ensuremath{(\mathrm{Ours})}}    & \underbar{78.6} & \underbar{61.6} & \textbf{59.5} & \underbar{1491.3} & \textbf{67.8} & \underbar{87.1} & \textbf{71.5} & \textbf{98.9\%} \\
\hline
\rowcolor{gray!20}
\multicolumn{9}{c}{\it{Retain 160 Tokens \textcolor{green!60!black}{($\downarrow$ 94.4\%)}}}\\
\hline
DivPrune {\scriptsize\ensuremath{(\mathrm{CVPR25})}}     & 75.0 & 59.3 & 54.1 & 1356.6 & 67.1 & 80.0 & 67.2 & 93.0\% \\
CDPruner {\scriptsize\ensuremath{(\mathrm{NeurIPS25})}}  & \underbar{76.8} & \underbar{60.8} & 55.4 & 1425.3 & 67.5 & \textbf{86.8} & \underbar{69.8} & \underbar{96.3\%} \\ 
EADP {\scriptsize\ensuremath{(\mathrm{ECCV26})}}
& \textbf{77.0} & \textbf{61.6} &  54.2 &  1419.5
& 67.4 &  86.0 & 69.5 & 96.0\% \\
\midrule
\textbf{CoRePruner} {\scriptsize\ensuremath{(\mathrm{Ours})}}    & \underbar{76.8} & 60.7 & \textbf{58.3} & 1445.7 & \textbf{67.9} & 86.4 & \textbf{70.4} & \textbf{97.4\%} \\
\textbf{CoRePruner$^{\dagger}$} {\scriptsize\ensuremath{(\mathrm{Ours})}}    & \underbar{76.8} & 60.5 & \underbar{58.2} & \underbar{1448.9} & \textbf{67.9} & \underbar{86.5} & \textbf{70.4} & \textbf{97.4\%} \\
\textbf{CoRePruner$^{\ddagger}$} {\scriptsize\ensuremath{(\mathrm{Ours})}}    & \underbar{76.8} & 60.6 & 58.1 & \textbf{1459.1} & \underbar{67.7} & \underbar{86.5} & \textbf{70.4} & \textbf{97.4\%} \\
\hline
\end{tabular*}
\caption{
Performance comparison on LLaVA-NeXT-7B under identical token budgets. CoRePruner consistently improves performance, demonstrating that our method generalizes beyond the LLaVA-1.5 backbone. Notation follows Table~\ref{tab:llava1.5}.}
\label{tab: llava-next}
\end{table*}

\subsection{Complete CoRePruner Procedure} 
\label{sec:complete-corepruner}
Algorithm~\ref{alg:corepruner} summarizes the complete procedure described
in Section~4. Stage~I constructs an
instruction-aware diverse subset $S$ of size $m_1$, and Stage II augments
it with $m_2$ reliable anchor tokens selected from the
remaining token pool. The final set $S^*=S\cup A$ therefore combines
task-relevant coverage with stable salient evidence.

\begin{table*}[t]
\centering
\small
\setlength{\tabcolsep}{3pt}

\begin{subtable}[t]{0.48\textwidth}
\centering
\begin{tabular*}{0.80\linewidth}{@{\extracolsep{\fill}} c | c c}

\toprule
\textbf{Retained Tokens} &
\textbf{Attention} &
\textbf{Random} \\
\midrule

\makebox[2em][r]{576}\;
\makebox[5em][l]{\textit{\textcolor{green!60!black}{(100\%)}}}
& \multicolumn{2}{c}{61.9} \\
\midrule

\makebox[2em][r]{448}\;
\makebox[5em][l]{\textit{\textcolor{green!60!black}{($\downarrow$22.2\%)}}}
& 58.5 & 61.4 \\

\makebox[2em][r]{320}\;
\makebox[5em][l]{\textit{\textcolor{green!60!black}{($\downarrow$44.4\%)}}}
& 54.1 & 60.6 \\

\makebox[2em][r]{192}\;
\makebox[5em][l]{\textit{\textcolor{green!60!black}{($\downarrow$66.7\%)}}}
& 50.4 & 60.1 \\

\makebox[2em][r]{64}\;
\makebox[5em][l]{\textit{\textcolor{green!60!black}{($\downarrow$88.9\%)}}}
& 38.7 & 55.2 \\

\bottomrule
\end{tabular*}
\caption{Clean condition}
\label{tab:clean_attn_vs_rand}
\end{subtable}
\hfill
\begin{subtable}[t]{0.48\textwidth}
\centering
\begin{tabular*}{0.80\linewidth}{@{\extracolsep{\fill}} c | c c}

\toprule
\textbf{Retained Tokens} &
\textbf{Attention} &
\textbf{Random} \\
\midrule

\makebox[2em][r]{576}\;
\makebox[5em][l]{\textit{\textcolor{green!60!black}{(100\%)}}}
& \multicolumn{2}{c}{62.3} \\
\midrule

\makebox[2em][r]{448}\;
\makebox[5em][l]{\textit{\textcolor{green!60!black}{($\downarrow$22.2\%)}}}
& 59.3 & 61.4 \\

\makebox[2em][r]{320}\;
\makebox[5em][l]{\textit{\textcolor{green!60!black}{($\downarrow$44.4\%)}}}
& 55.8 & 63.2 \\

\makebox[2em][r]{192}\;
\makebox[5em][l]{\textit{\textcolor{green!60!black}{($\downarrow$66.7\%)}}}
& 53.0 & 60.1 \\

\makebox[2em][r]{64}\;
\makebox[5em][l]{\textit{\textcolor{green!60!black}{($\downarrow$88.9\%)}}}
& 41.7 & 55.1 \\

\bottomrule
\end{tabular*}
\caption{Perturbed condition}
\label{tab:pert_attn_vs_rand}
\end{subtable}

\caption{
Performance (\%) on GQA after removing tokens selected by attention vs. random pruning.
(a) On clean images, removing the top-$k$ attended tokens ($T_k$) causes a larger accuracy drop than removing randomly selected tokens.
(b) On perturbed images, we remove $T_k$ identified from clean inputs. Removing $T_k$ still results in a substantially larger performance drop than random pruning.
}
\label{tab:robustness}
\end{table*}

\section{Additional Experiments}

\subsection{Image Understanding on LLaVA-NeXT-7B.}
\label{sec:llava-next-results}
We further evaluate CoRePruner on LLaVA-NeXT-7B at token budgets of
640, 320, and 160.
We compare against representative attention-based, diversity-based, and
hybrid pruning methods. The three CoRePruner variants use cosine, RBF,
and polynomial kernels for the Stage-I similarity function, following
Table~\ref{tab:llava1.5}.
As shown in Table~\ref{tab: llava-next}, CoRePruner achieves the highest
average performance and relative retention across all three token budgets,
retaining 100.0\%, 98.9\%, and 97.4\% of the full-token performance,
respectively.
The consistent gains across token budgets and kernel choices indicate that
the improvements arise from reliability-aware complementary token
construction rather than a particular Stage-I kernel, further supporting
the architectural generalizability of CoRePruner.

\subsection{Video Understanding.}
To further evaluate the effectiveness of our method, we conduct additional experiments on video question-answering tasks using Qwen2.5-VL-7B. Table~\ref{tab:video} reports the performance on the MVBench and VideoMME benchmarks under different token retention budgets.
CoRePruner matches or outperforms VisPruner and CDPruner on most metrics, with clearer gains at 85\% and 90\% token reduction. At a more aggressive 95\% reduction (314 of 6,272 tokens), CoRePruner obtains 55.7 on MVBench, compared with 54.2 for VisPruner and 55.8 for CDPruner. These results suggest that the proposed construction remains applicable to longer visual sequences, although the video evaluation is limited to a single backbone and two benchmarks.

\begin{table}[!t]
\centering
\small
\setlength{\tabcolsep}{3pt}
\begin{tabular*}{0.80\columnwidth}
{@{\extracolsep{\fill}} c | c c}
\toprule
\textbf{Retained Tokens}
& \textbf{Ratio}
& \textbf{Recall@$k$} \\
\midrule

\makebox[2em][r]{448}\,
\makebox[4.8em][l]{\textit{\textcolor{green!60!black}{($\downarrow$22.2\%)}}}
& 3.13 & 0.91 \\

\makebox[2em][r]{320}\,
\makebox[4.8em][l]{\textit{\textcolor{green!60!black}{($\downarrow$44.4\%)}}}
& 1.99 & 0.78 \\

\makebox[2em][r]{192}\,
\makebox[4.8em][l]{\textit{\textcolor{green!60!black}{($\downarrow$66.7\%)}}}
& 1.78 & 0.82 \\

\makebox[2em][r]{64}\,
\makebox[4.8em][l]{\textit{\textcolor{green!60!black}{($\downarrow$88.9\%)}}}
& 2.52 & 0.70 \\

\bottomrule
\end{tabular*}
\caption{
Quantitative analysis of token prominence under perturbations.
The high \textit{Ratio} (up to 3.13) and \textit{Recall@$k$} (up to 0.91) show that tokens identified as important under clean inputs remain prominent and are largely preserved within the perturbed top-$k$ set.
}
\label{tab:attn_recall}
\end{table}

\begin{figure}[t]
    \centering
    \includegraphics[width=1.0\linewidth]{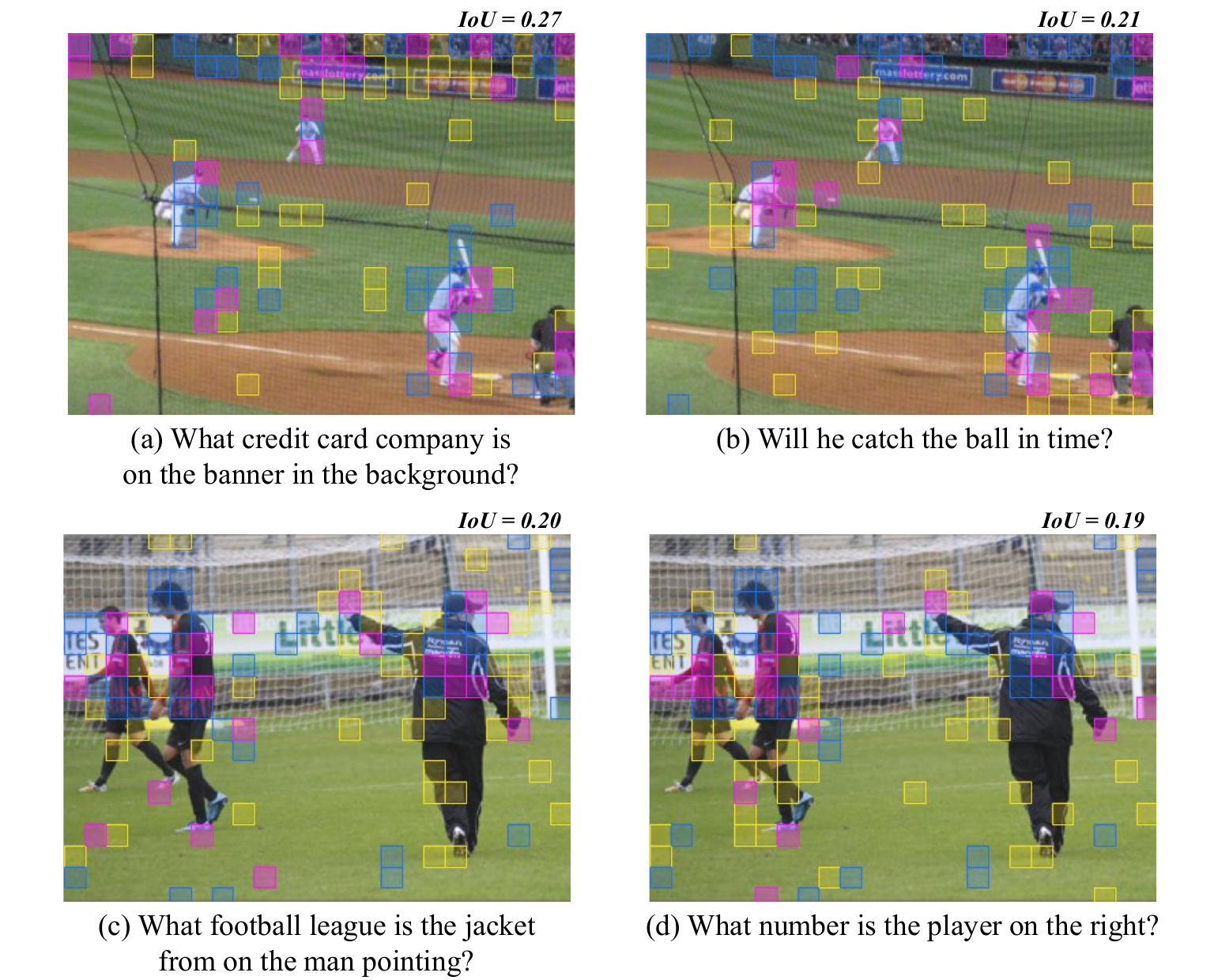}
    \caption{
    Comparison of token selection for the same image under different instructions.
    \textcolor{Goldenrod}{Instruction-aware diversity-based token selection (yellow)} shifts toward instructions-relevant regions, while \textcolor{RoyalBlue}{attention-based token selection (blue)} focuses on visually salient areas. \textcolor{VioletRed}{Overlap} is shown in \textcolor{VioletRed}{pink}.
    }
    \label{fig:qual_question_relevance}
\end{figure}

\section{Additional Analysis}
\subsection{Robustness of Semantically Critical Tokens}


\begin{figure*}[!ht]
    \centering
    \includegraphics[width=1.0\linewidth]{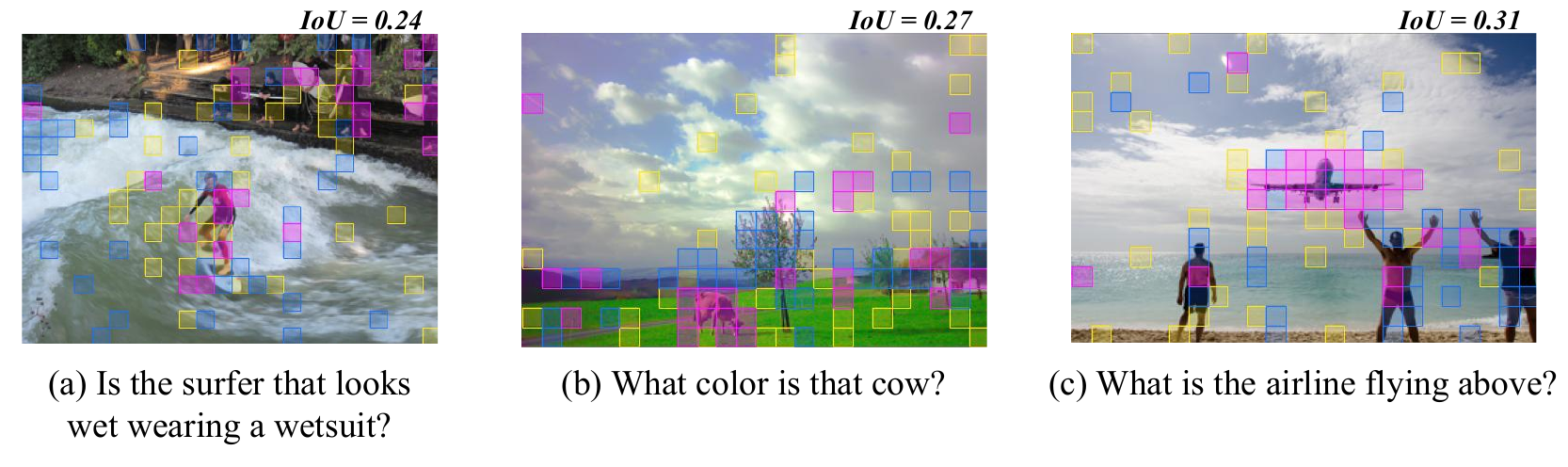}
    \caption{
    High-overlap cases between diversity-based and attention-based token selections.
    When questions target prominent foreground objects, both strategies converge on the queried instance, leading to relatively high IoU. Notation follows Figure~\ref{fig:qual_question_relevance}.
    }
    \label{fig:qual_high_iou}
\end{figure*}

\begin{figure*}[!ht]
    \centering
    \includegraphics[width=1.0\linewidth]{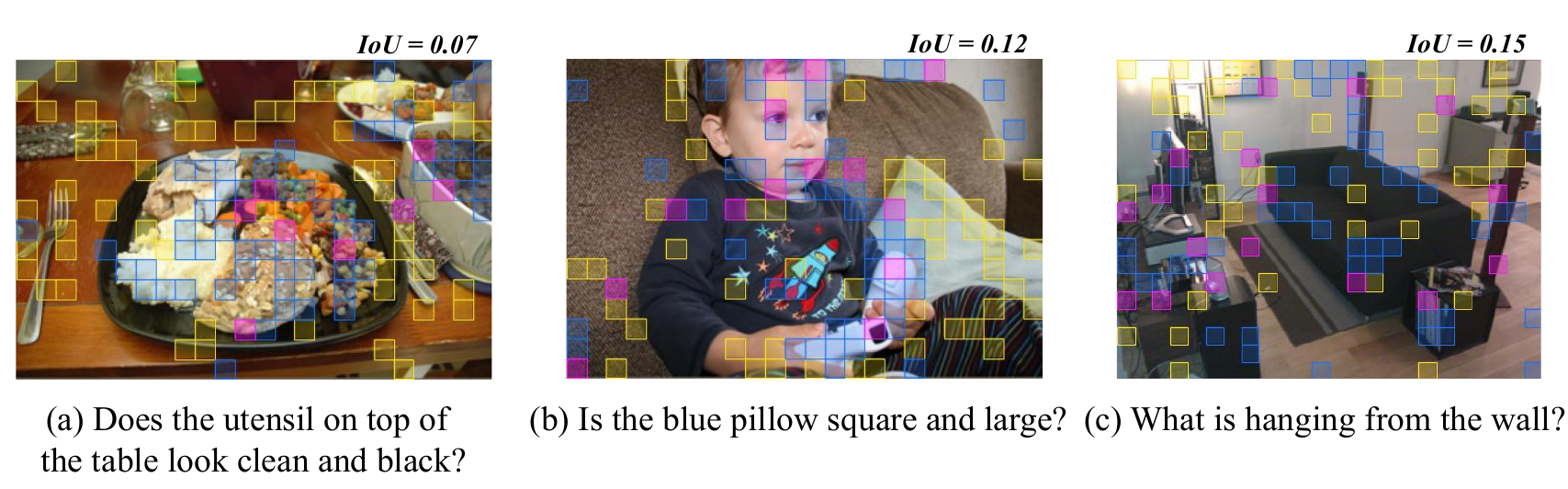}
    \caption{
    Low-overlap cases between diversity-based and attention-based token selections.
    For questions targeting background or peripheral cues, attention-based selection focuses on salient regions while instruction-aware diversity captures subtle task-relevant evidence. Notation follows Figure~\ref{fig:qual_question_relevance}.
    }
    \label{fig:qual_low_iou}
\end{figure*}

To verify that semantically critical tokens remain robust under input perturbations, we conduct additional analyses under both clean and perturbed conditions.

First, we analyze the performance drop caused by removing tokens selected by attention-based pruning. Table~\ref{tab:robustness}(a) reports the results on clean inputs. Removing the top-$k$ attended tokens ($T_k$) consistently leads to a substantially larger accuracy drop on the GQA benchmark than removing the same number of randomly selected tokens. This observation indicates that tokens receiving high attention under clean inputs correspond to semantically important visual evidence for downstream reasoning.

Next, we examine whether these semantically important tokens remain critical under perturbations. In Table~\ref{tab:robustness}(b), we remove the same token set $T_k$ identified from clean inputs but evaluate the model on perturbed images. Even under perturbation, removing these tokens still causes a significantly larger performance drop than random pruning. This result suggests that tokens identified as important under clean conditions remain functionally critical even when the input is perturbed.

To further analyze this phenomenon, we examine whether tokens that receive high attention under clean inputs remain prominent under perturbations. We employ two metrics: \textit{Ratio} and \textit{Recall@k}. The \textit{Ratio} measures how much higher the attention scores of the clean top-$k$ tokens ($T_k$) remain compared to the remaining tokens ($T_k^c$) under perturbation:
\begin{equation}
\text{Ratio} =
\frac{\mu(A' \mid T_k)}{\mu(A' \mid T_k^c)},
\end{equation}
where $A'$ denotes the attention scores under perturbation and $\mu(A' \mid S)$ represents the mean attention score over token set $S$.

We also compute \textit{Recall@k}, which measures the fraction of tokens that remain within the top-$k$ attention ranks after perturbation:
\begin{equation}
\text{Recall@}k =
\frac{|TopK(\text{pert}) \cap TopK(\text{clean})|}{k}.
\end{equation}

Table~\ref{tab:attn_recall} shows that tokens identified as important under clean inputs continue to receive substantially higher attention than the remaining tokens even under perturbation. The attention ratio between $T_k$ and $T_k^c$ ranges from $1.78\times$ to $3.13\times$, while Recall@k remains between $0.70$ and $0.91$ across different token budgets. These results indicate that tokens identified as important under clean inputs remain consistently prominent and largely preserved within the perturbed top-$k$ set.

Overall, these findings suggest that although attention scores themselves may fluctuate under perturbations, the tokens receiving high attention correspond to semantically critical visual evidence whose functional importance persists even when the input is perturbed.

\subsection{Qualitative Comparison}
To further validate the observations in Section 3, we provide additional qualitative comparisons between token subsets selected by instruction-aware diversity-based and attention-based methods. These examples demonstrate that while both strategies preserve essential evidence for answering questions, they often identify substantially different sets of informative tokens depending on the query context. We further analyze these cases to provide additional insights into the complementary roles of different token selection strategies.

\begin{table*}[!t]
\centering
\footnotesize
\setlength{\tabcolsep}{2.5pt}
\renewcommand{\arraystretch}{1.05}

\begin{tabular*}{\textwidth}{
@{\extracolsep{\fill}}
l | c c c c c c c c c | c
}
\toprule
\textbf{Method}
& \textbf{VQA}\textsuperscript{v2}
& \textbf{GQA}
& \textbf{VQA}\textsuperscript{Text}
& \textbf{MME}
& \textbf{MMB}\textsuperscript{EN}
& \textbf{MMB}\textsuperscript{CN}
& \textbf{MMVet}
& \textbf{SQA}\textsuperscript{Img}
& \textbf{POPE}
& \textbf{Avg.}
\\
\midrule

\rowcolor{gray!20}
\multicolumn{11}{c}{
\itshape
Retain 128 Tokens
\textcolor{green!60!black}{($\downarrow$ 77.8\%)}
}
\\
\midrule

DivPrune
& 76.0 & 59.4 & 55.9 & 1405.1
& 61.5 & 54.8 & 30.6 & 68.6 & 87.0 & 62.7
\\

VisPruner
& 75.8 & 58.2 & 57.0 & 1461.4
& 62.7 & 57.3 & 33.7 & 69.1 & 84.6 & 63.5
\\

CDPruner
& 76.6 & 59.9 & 56.2 & 1431.4
& 63.1 & 55.0 & 32.8 & 69.0 & 87.7 & 63.5
\\
\midrule

\textbf{CoRePruner}
& $77.0_{\pm 0.02}$
& $60.2_{\pm 0.05}$
& $57.2_{\pm 0.08}$
& $1457.6_{\pm 1.04}$
& $63.2_{\pm 0.10}$
& $56.5_{\pm 0.16}$
& $33.3_{\pm 0.36}$
& $68.9_{\pm 0.16}$
& $87.6_{\pm 0.03}$
& 64.1
\\

\textbf{CoRePruner$^{\dagger}$}
& $76.9_{\pm 0.02}$
& $60.3_{\pm 0.06}$
& $57.3_{\pm 0.05}$
& $1449.8_{\pm 1.64}$
& $62.9_{\pm 0.09}$
& $56.9_{\pm 0.26}$
& $32.7_{\pm 0.25}$
& $68.6_{\pm 0.10}$
& $87.5_{\pm 0.09}$
& 64.0
\\

\textbf{CoRePruner$^{\ddagger}$}
& $76.9_{\pm 0.01}$
& $60.4_{\pm 0.03}$
& $57.5_{\pm 0.18}$
& $1449.9_{\pm 1.24}$
& $63.0_{\pm 0.09}$
& $57.2_{\pm 0.21}$
& $32.9_{\pm 0.23}$
& $68.5_{\pm 0.12}$
& $87.6_{\pm 0.05}$
& 64.1
\\
\midrule

\rowcolor{gray!20}
\multicolumn{11}{c}{
\itshape
Retain 64 Tokens
\textcolor{green!60!black}{($\downarrow$ 88.9\%)}
}
\\
\midrule

DivPrune
& 74.1 & 57.5 & 54.5 & 1334.7
& 60.1 & 52.3 & 28.1 & 68.0 & 85.8 & 60.8
\\

VisPruner
& 72.7 & 55.4 & 55.8 & 1369.9
& 61.3 & 55.1 & 32.3 & 69.1 & 80.4 & 61.2
\\

CDPruner
& 75.4 & 58.6 & 55.3 & 1415.1
& 61.1 & 53.2 & 30.5 & 68.1 & 87.5 & 62.3
\\
\midrule

\textbf{CoRePruner}
& $75.8_{\pm 0.01}$
& $59.0_{\pm 0.03}$
& $56.5_{\pm 0.16}$
& $1435.8_{\pm 1.95}$
& $61.5_{\pm 0.16}$
& $56.2_{\pm 0.41}$
& $30.9_{\pm 0.41}$
& $69.0_{\pm 0.05}$
& $87.6_{\pm 0.02}$
& 63.1
\\

\textbf{CoRePruner$^{\dagger}$}
& $75.9_{\pm 0.01}$
& $60.2_{\pm 0.02}$
& $56.4_{\pm 0.05}$
& $1438.0_{\pm 2.38}$
& $61.6_{\pm 0.20}$
& $56.0_{\pm 0.48}$
& $31.7_{\pm 0.15}$
& $68.9_{\pm 0.10}$
& $87.5_{\pm 0.15}$
& 63.3
\\

\textbf{CoRePruner$^{\ddagger}$}
& $75.8_{\pm 0.01}$
& $58.9_{\pm 0.08}$
& $55.0_{\pm 0.05}$
& $1443.2_{\pm 1.83}$
& $61.5_{\pm 0.14}$
& $55.9_{\pm 0.44}$
& $30.7_{\pm 0.17}$
& $69.5_{\pm 0.18}$
& $87.9_{\pm 0.05}$
& 63.0
\\
\bottomrule
\end{tabular*}
\caption{
Performance of LLaVA-1.5-7B under 128- and 64-token budgets.
CoRePruner results are reported as mean$\pm$standard deviation over five seeds. Notation follows Table~\ref{tab:llava1.5}.
}
\label{tab:five_seed_results}
\end{table*}

\begin{table*}[!t]
\centering
\footnotesize
\setlength{\tabcolsep}{7pt}
\renewcommand{\arraystretch}{1.05}

\begin{tabular*}{\textwidth}{@{\extracolsep{\fill}}l | c c c | c c}
\toprule
\textbf{Benchmark}
&
\textbf{Stage I-only}
&
\textbf{Stage II-only}
&
\textbf{w/o $\rho$}
&
\textbf{$\rho=\exp(-u)$}
&
\textbf{Ours ($\rho=\frac{1}{1+u}$)}
\\
\midrule

\rowcolor{gray!20}
\multicolumn{6}{c}{
\itshape
Retain 128 Tokens
\textcolor{green!60!black}{($\downarrow$ 77.8\%)}
}
\\
\midrule

VQA\textsuperscript{Text}
& $56.2_{\pm 0.02}$
& $57.0_{\pm 0.08}$
& $57.0_{\pm 0.05}$
& $57.2_{\pm 0.08}$
& $57.2_{\pm 0.12}$
\\

GQA
& $59.7_{\pm 0.02}$
& $58.7_{\pm 0.07}$
& $59.9_{\pm 0.04}$
& $60.2_{\pm 0.05}$
& $60.2_{\pm 0.03}$
\\

MME
& $1419.4_{\pm 0.87}$
& $1415.6_{\pm 4.68}$
& $1449.7_{\pm 2.37}$
& $1457.6_{\pm 1.04}$
& $1455.8_{\pm 1.80}$
\\

MMB\textsuperscript{EN}
& $62.7_{\pm 0.19}$
& $62.0_{\pm 0.35}$
& $62.5_{\pm 0.25}$
& $63.2_{\pm 0.10}$
& $63.5_{\pm 0.12}$
\\

MMVet
& $30.6_{\pm 0.78}$
& $31.5_{\pm 0.26}$
& $32.4_{\pm 0.19}$
& $33.3_{\pm 0.36}$
& $33.2_{\pm 0.39}$
\\
\midrule

\rowcolor{gray!20}
\multicolumn{6}{c}{
\itshape
Retain 64 Tokens
\textcolor{green!60!black}{($\downarrow$ 88.9\%)}
}
\\
\midrule

VQA\textsuperscript{Text}
& $55.1_{\pm 0.02}$
& $55.9_{\pm 0.08}$
& $56.1_{\pm 0.02}$
& $56.5_{\pm 0.06}$
& $56.5_{\pm 0.16}$
\\

GQA
& $58.7_{\pm 0.00}$
& $56.1_{\pm 0.18}$
& $58.6_{\pm 0.00}$
& $59.0_{\pm 0.08}$
& $59.0_{\pm 0.03}$
\\

MME
& $1404.9_{\pm 2.66}$
& $1430.5_{\pm 5.51}$
& $1426.9_{\pm 4.84}$
& $1437.5_{\pm 2.18}$
& $1435.8_{\pm 1.95}$
\\

MMB\textsuperscript{EN}
& $60.9_{\pm 0.14}$
& $60.8_{\pm 0.23}$
& $61.1_{\pm 0.07}$
& $61.3_{\pm 0.10}$
& $61.4_{\pm 0.16}$
\\

MMVet
& $26.7_{\pm 0.77}$
& $30.3_{\pm 0.91}$
& $30.5_{\pm 0.07}$
& $31.0_{\pm 0.58}$
& $31.1_{\pm 0.41}$
\\

\bottomrule
\end{tabular*}

\caption{
Stage-wise and $\rho$-form ablation on LLaVA-1.5-7B under 128- and 64-token budgets.
Results are reported as mean$\pm$standard deviation over five seeds.
}
\label{tab:stage_rho_ablation}

\end{table*}



\subsubsection{Instruction-Aware Diversity Reflects Question Relevance.} 
First, we present cases showing how the instruction-aware diversity-based method dynamically adapts token selection to the input question, even for the same image.
As shown in Figure~\ref{fig:qual_question_relevance}(a) and (b) from VQAv2, when the question asks, \emph{``What credit card company is on the banner in the background?''}, the diversity-selected tokens are concentrated on the background banner. In contrast, for the question \emph{``Will he catch the ball in time?''}, the selection shifts towards the foreground action.
Similarly, in Figure~\ref{fig:qual_question_relevance}(c) and (d) from TextVQA, the method focuses on the \emph{pointing man} or the \emph{player on the right} depending on which subject is specified in the prompt.
These results highlight that instruction-aware diversity is highly sensitive to the question's focus, rather than merely relying on static, visually salient regions.

\subsubsection{High Overlap in Object-Centric Questions.}
We next analyze cases where diversity-based and attention-based selections exhibit high overlap. The overlap between the two strategies tends to increase when a question directly targets a prominent foreground object. In such instances, both strategies naturally converge on the tokens representing the queried instance, leading to a relatively high Intersection-over-Union (IoU).
As illustrated in Figure~\ref{fig:qual_high_iou}, for questions focusing on distinct objects such as (a) \emph{surfer}, (b) \emph{cow}, and (c) \emph{airplane}, the overlapping tokens are densely clustered around these subjects, with IoU values ranging from 0.24 to 0.31.

\subsubsection{Low Overlap in Background-Related Questions.}
Conversely, the overlap significantly decreases when the question pertains to background elements or non-central contextual cues. For example, in Figure~\ref{fig:qual_low_iou}, questions such as \emph{``What is hanging from the wall?''} or \emph{``What is the name of the cooking utensil that is hanging from the hook?''} require focusing on peripheral regions. In these cases, the IoU drops to 0.07--0.15. This divergence indicates that while attention-based pruning often remains anchored to globally salient regions, the instruction-aware diversity method captures spatially subtle but task-relevant evidence.

Overall, these qualitative findings reinforce our claim that informative token selection is not unique. While attention-based pruning favors general visual saliency, instruction-aware diversity-based pruning better tracks the specific semantic target of the instruction. This functional complementarity justifies our approach of integrating both strategies for more robust token construction.

\subsection{Seed Sensitivity}
Tables~\ref{tab:five_seed_results} and~\ref{tab:stage_rho_ablation} report the 128- and 64-token results of Tables~\ref{tab:llava1.5} and~\ref{tab:ablation-study} with mean$\pm$standard deviation over five seeds. Stochasticity arises from the perturbation patterns, which differ across seeds; the baselines are deterministic and are reported from a single run. CoRePruner's standard deviations are at most $0.41$ on percentage metrics and $1.95$ MME points, and the ablation gaps generally exceed them, so the reported gains are not an artifact of perturbation sampling.

\end{document}